\documentclass[10pt,twocolumn,letterpaper]{article}
\usepackage{xr-hyper}
\usepackage[accsupp]{axessibility} 
\usepackage[pagenumbers]{cvpr} 

\usepackage[dvipsnames]{xcolor}

\definecolor{cvprblue}{rgb}{0.21,0.49,0.74}
\usepackage[pagebackref,breaklinks,colorlinks,citecolor=cvprblue]{hyperref}
\newcommand{\tabincell}[2]{\begin{tabular}{@{}#1@{}}#2\end{tabular}}
\newcommand{\rightcomment}[1]{\hfill\textcolor{gray}{\% #1}}
\newcommand{\rightcommentblue}[1]{\hfill\textcolor{blue}{\% #1}}
\usepackage{forloop}
\usepackage{makecell}

\usepackage{color}

\usepackage{hyperref}
\usepackage{url}
\usepackage{algorithm}
\usepackage[noend]{algpseudocode}
\usepackage{caption}
\usepackage{subcaption}
\usepackage{mathtools}
\usepackage{multicol}
\usepackage{multirow}
\usepackage{makecell}
\usepackage{bbding}
\definecolor{minetable1colorx}{rgb}{0.75, 0.75, 0.75}

\newcommand{\mineyes}{{\scriptsize \CheckmarkBold}}
\newcommand{\mineno}{{\scriptsize \textcolor{minetable1colorx}{\XSolidBrush}}}
\definecolor{minetable1colorx}{rgb}{0.75, 0.75, 0.75}
\usepackage{booktabs}       
\usepackage{bbm}
\usepackage{caption}
\usepackage{graphicx}

\usepackage{xcolor,colortbl}
\usepackage[misc]{ifsym}
\newcounter{ct}
\newcommand{\markdent}[1]{\forloop{ct}{0}{\value{ct} < #1}{\hspace{\algorithmicindent}}}
\newcommand{\markcomment}[1]{\Statex\markdent{#1}}
\definecolor{lightgrey}{gray}{0.9}

\title{SmartRefine: A Scenario-Adaptive Refinement Framework\\ for Efficient Motion Prediction}

\author{%
Yang Zhou$^{1*}$~~~~~~ Hao Shao$^{1, 2*}$~~~~~~ Letian Wang$^{3}$ \\ \vspace{0.5em}Steven L. Waslander$^{3}$~~ Hongsheng Li$^{2,4,5}$ ~~ Yu Liu$^{1,5}$~\textsuperscript{\Letter} \\ 
$^1$SenseTime Research~~~ $^2$CUHK MMLab ~~~ $^3$University of Toronto\\ ~~~ $^4$CPII under InnoHK ~~~ $^5$Shanghai Artificial Intelligence Laboratory
}

\begin{document}

\maketitle

\makeatletter{\renewcommand*{\@makefnmark}{}
\footnotetext{$^{*}$ Equal contribution, \textsuperscript{\Letter} Corresponding author.}
\footnotetext{$^{1}$ November 2023.}
\makeatother}

\begin{abstract}
Predicting the future motion of surrounding agents is essential for autonomous vehicles (AVs) to operate safely in dynamic, human-robot-mixed environments. Context information, such as road maps and surrounding agents' states, provides crucial geometric and semantic information for motion behavior prediction. To this end, recent works explore two-stage prediction frameworks where coarse trajectories are first proposed, and then used to select critical context information for trajectory refinement. However, they either incur a large amount of computation or bring limited improvement, if not both. In this paper, we introduce a novel scenario-adaptive refinement strategy, named SmartRefine, to refine prediction with minimal additional computation. Specifically, SmartRefine can comprehensively adapt refinement configurations based on each scenario's properties, and smartly chooses the number of refinement iterations by introducing a quality score to measure the prediction quality and remaining refinement potential of each scenario. SmartRefine is designed as a generic and flexible approach that can be seamlessly integrated into most state-of-the-art motion prediction models. Experiments on Argoverse (1 \& 2) show that our method consistently improves the prediction accuracy of multiple state-of-the-art prediction models. 
Specifically, by adding SmartRefine to QCNet, we outperform all published ensemble-free works on the Argoverse 2 leaderboard (single agent track) at submission$^1$. 
Comprehensive studies are also conducted to ablate design choices and explore the mechanism behind multi-iteration refinement. Codes are available at our \href{https://github.com/opendilab/SmartRefine/}{webpage}.
\end{abstract}

\vspace{-1em}

\section{Introduction}

Predicting the future motion of surrounding agents (\textit{e.g.}, vehicle, cyclist, pedestrian) is crucial for autonomous driving frameworks~\cite{hu2023planning, shao2023lmdrive, shao2023reasonnet, shao2023safety,wang2023efficient} to safely and efficiently make decisions in a dynamic and human-robot-mixed environment.
Context information, such as high-definition maps (HD maps) and surrounding agents' states, provides crucial geometric and semantic information for motion behavior, as agents' behaviors are highly dependent on the map topology and impacted by interaction with surrounding agents. For instance, vehicles usually move in drivable areas and follow the direction of lanes, and agents' interactive cues such as yielding would inform other agent's decision-making. 
As a result, recent motion prediction models are shown to significantly benefit from delicate context representation designs~\citep{chai2019multipath,gao2020vectornet,liang2020learning,wang2022transferable} and context encodings~\citep{salzmann2020trajectron++,ngiam2021scene,varadarajan2022multipath++}. However, complicated context encodings usually come at a high computational cost and high memory footprint. Since vehicles are high-speed robots and minimal delay could result in catastrophic accidents, recent advanced state-of-the-art methods could have low applicability due to limitations in mobile computation capacity and the hard real-time requirements of autonomous driving. 

By contrast, human drivers can easily predict surrounding agents' future behaviors, even if they confront a daunting amount of context information. 
As implied by neuroscience, humans’ efficient reasoning capabilities benefit from their \textit{selective attention} mechanism~\citep{niv2019learning,radulescu2019holistic}, which identifies compact context information critical to the task for efficient reasoning. Similarly, motion prediction models are shown to be able to produce high-quality predictions with only a few critical context elements provided, such as only giving the ground-truth future reference lanes and conducting prediction in the relative coordinates~\citep{wang2021hierarchical,ye2023improving,wang2021socially}. Therefore, if we can identify the critical context elements, and aggregate more information from these critical inputs to further refine the predictions, both the computational efficiency and prediction performance can be significantly improved. 

While multiple refinement strategies have been proposed, the proper design of refinement is non-trivial. \citep{zhou2023query,ye2023bootstrap} apply a prediction backbone to generate trajectory proposals, which are then used as anchor trajectories for refinement. While accuracy improvement is observed, the refinement still employs all context information without selection, which leads to high computation costs. \citep{choi2022r,shi2022motion} propose pooling and grouping methods to select context features via fixed manual rules and refine trajectory for fixed iterations. 
However, the mechanism of refinement is still under-studied, and only fixed refinement strategies are applied, where many iterations of refinement usually exchange a large amount of computation for limited performance improvement. 

To this end, we therefore propose a scenario-adaptive refinement strategy, SmartRefine, which improves prediction accuracy with minimal additional computation for a wide variety of prediction models. Our key insight is that, while motion prediction models confront various driving scenarios, the prediction quality and refinement potential in different scenarios are not uniform: though some scenarios can benefit from intense refinement, some scenarios can be insensitive to refinement or even pushed away from ground truth due to over-refinement. Further, different scenarios may require different refinement configurations (\textit{i.e.} how to select and encode context), while previous methods typically apply fixed refinement configurations to all scenarios (as shown in Table~\ref{tab:refinement summary}). 
To address these issues, the proposed SmartRefine can adapt the anchor/context selection and context encoding according to each scenario's properties, and can select the number of refinement iterations by introducing a quality score to measure the prediction quality and remaining refinement potential of each scenario.
Thus, our method can allocate computation resources to those scenarios that require refinement, and terminate the refinement of the other scenarios to avoid worse prediction and wasted computation, achieving a better trade-off between accuracy and efficiency.

It is worth noting that SmartRefine is not only lightweight, but also designed as a general and flexible framework that is decoupled from the primary prediction model backbone, and only requires a generic interface to the model backbone (predicted trajectories and trajectory features).
Therefore, SmartRefine can be easily integrated into most state-of-the-art motion prediction models. This distinguishes our method from previous refinement methods~\citep{zhou2023query,ye2023bootstrap,choi2022r,shi2022motion} which are either computationally heavy, or highly coupled with a particular backbone, if not both.

To summarize, the contributions of this work are threefold:
\begin{itemize}
    \item We introduce SmartRefine, a scenario-adaptive refinement method that considers comprehensive design choices and configurations for refinement and adapts them to each scenario, to effectively enhance prediction accuracy with limited additional computation.
    \item We propose a generic and flexible refinement framework, which can be easily integrated into most state-of-the-art motion prediction models to enhance performance. Codes are released to facilitate further research.
    \item We conduct extensive experiments on Argoverse and Argoverse 2 datasets (both the validation set and test set), where we evaluate SmartRefine by applying it to multiple state-of-the-art motion prediction models. We show that SmartRefine improves the accuracy of all considered motion prediction models with little additional computation. Specifically, by adding SmartRefine to QCNet~\cite{zhou2023query}, we outperform all published ensemble-free works on the Argoverse 2 leaderboard (single agent track) at the time of the paper submission. 
\end{itemize}

\begin{table}[t!]
	\centering
	\vspace{5pt}
		\resizebox{1.0\linewidth}{!}{
		\begin{tabular}{lccccccc}
			\toprule
   & {\multirow{3}{*}{\makecell[c]{Refine}}}& \multicolumn{3}{c}{\makecell[c]{Context Selection/Encoding}}&\multicolumn{3}{c}{\makecell[c]{Refinement Iteration}}
   \\
   \cmidrule(lr){3-5} 
   \cmidrule(lr){6-8} 
 & 
   \multicolumn{1}{c}{ }&
   \multirow{2}{*}{\makecell[c]{All}} & \multirow{2}{*}{\makecell[c]{Fixed\\Strategy}} & \multirow{2}{*}{\makecell[c]{Adaptive\\Strategy}}
   &
     \multirow{2}{*}{\makecell[c]{Single}} & \multirow{2}{*}{\makecell[c]{Multiple}} & \multirow{2}{*}{\makecell[c]{Scenario\\Adaptive}}
   \\
   \\
			\midrule
            TNT \citep{zhao2021tnt}  
			 & \mineno \\
            GoalNet \citep{zhang2021map}  
		   & \mineno  \\
            GANet \citep{wang2023ganet} 
              & \mineno \\
            ProphNet \citep{wang2023prophnet}
              & \mineno\\
                        \midrule
            DCMS \citep{ye2023bootstrap} 
             & \mineyes & \mineno & \mineno & \mineno & \mineyes & \mineno & \mineno  \\
            QCNet \citep{zhou2023query}
		   & \mineyes & \mineyes & \mineno & \mineno & \mineyes & \mineno & \mineno \\
            R-Pred \citep{choi2022r}
             & \mineyes & \mineno & \mineyes & \mineno & \mineyes & \mineno & \mineno \\
            MTR \citep{shi2022motion}
			& \mineyes & \mineno & \mineyes & \mineno & \mineno & \mineyes & \mineno\\
			\toprule
			\textbf{SmartRefine (ours)}  
			& \mineyes & \mineno & \mineno & \mineyes & \mineno & \mineno & \mineyes  \\
			\bottomrule
		\end{tabular}
			}
   \caption{A comparison between the proposed SmartRefine framework and previous methods in terms of 1) whether refinement is conducted; 2) how the context selection and encoding is conducted; 3) how to determine the number of refinement iterations. 
   }
	\label{tab:refinement summary}
\end{table}

\section{Related Work}
\subsection{Goal-Conditioned Trajectory Prediction}
Trajectory prediction typically takes the road map, agent history states, and semantic type as inputs and outputs future agent states. To encode scene context (road map, surround agents) around the target agents, early works~\citep{cui2019multimodal, chai2019multipath} rasterize them into a bird-eye-view image and process it with convolutional neural networks. To encourage relationship reasoning and reduce computation, recent works adopt vector-based encoding schemes, where each scene context is represented as a vector, and encoded with permutation-invariant operators such as pooling~\citep{varadarajan2022multipath++}, graph convolution~\citep{da2022path}, and attention mechanism~\citep{wang2023prophnet,li2024scenarionet,feng2023trafficgen}.
Inspired by human hierarchical decision-making where the motion intention determines the specific trajectory, 
goal-conditional prediction, which first predicts or predefines goal candidates, and then predicts trajectory conditioned on them, is shown to be effective and has been widely adopted in state-of-the-art methods.
Multipath~\citep{chai2019multipath} predefines a set of anchor trajectories by clustering and predicts offset for the trajectories.
TNT~\citep{zhao2021tnt} predicts goal points that offset from a lane centerline.
GoalNet~\citep{zhang2021map} uses lane segments as trajectory anchors.
we predict the lane that a vehicle will pass in the future.
FRM~\citep{park2023leveraging} predicts the occupancy of each waypoint on the lanes, and then predicts the fine-grained trajectory. 
GANet~\citep{wang2023ganet} proposes a goal area-based framework for predicting goal areas and fusing crucial distant map features.
MTR~\citep{shi2022motion} pre-defines a set of goal points as queries by clustering the trajectory data of each agent type, and adopts attention layers to aggregate context information based on the queries.
Prophnet~\citep{wang2023prophnet} uses trajectory proposals as learnable anchors to enable the attention layers to encode goal-oriented scene contexts.
However, while these methods exploit goal-conditioned prediction, they only leverage goal-conditioned contexts once. In this paper, we utilize trajectory anchors as goals in an iterative manner, where the predicted trajectory at one iteration can benefit the next iteration by specifying more precise anchors as goals to extract more relevant map information.
\vspace{-0.25em}
\subsection{Refinement: Two-Stage Trajectory Prediction}
\vspace{-0.25em}
Inspired by refinement networks~\citep{ren2015faster, carion2020end} in computer vision, refinement strategies have been introduced to the trajectory prediction community. Trajectory refinement typically takes proposed trajectories in the first stage as inputs, and outputs the offset and probability for each proposed trajectory. 
DCMS~\citep{ye2023bootstrap} takes the output of the first stage as anchor trajectories and conducts refinement to predict the offset.
QCNet~\citep{zhou2023query} uses a small GRU to embed the proposed trajectories in the first stage and predict the offset of the trajectories by fusing the same scene context.
R-Pred~\citep{choi2022r} proposes tube-query scene attention layers to refine based on local contexts, and interaction attention layers to refine based on agents' interactions.
MTR~\citep{shi2022motion} uses the predicted trajectory as anchors to retrieve contexts along the trajectory for refinement. The refinement is conducted for multiple iterations.
However, existing methods typically employ a relatively large refinement network or utilize fixed refinement strategies (e.g. retrieval range, context encodings, and refinement iterations), which leads to sub-optimal performance in various scenarios. 
By contrast, SmartRefine can comprehensively adapt refinement configurations based on each scenario’s properties, and smartly choose the number of refinement iterations. Besides, SmartRefine is designed as a lightweight and flexible approach that can be seamlessly integrated into state-of-the-art motion prediction. We will show through experiments how our method can refine prediction with minimal additional computation.

\section{Method}
\begin{figure*}[!t]
  \centering
  \hspace*{-0.0\linewidth}\includegraphics[width=0.99\linewidth]{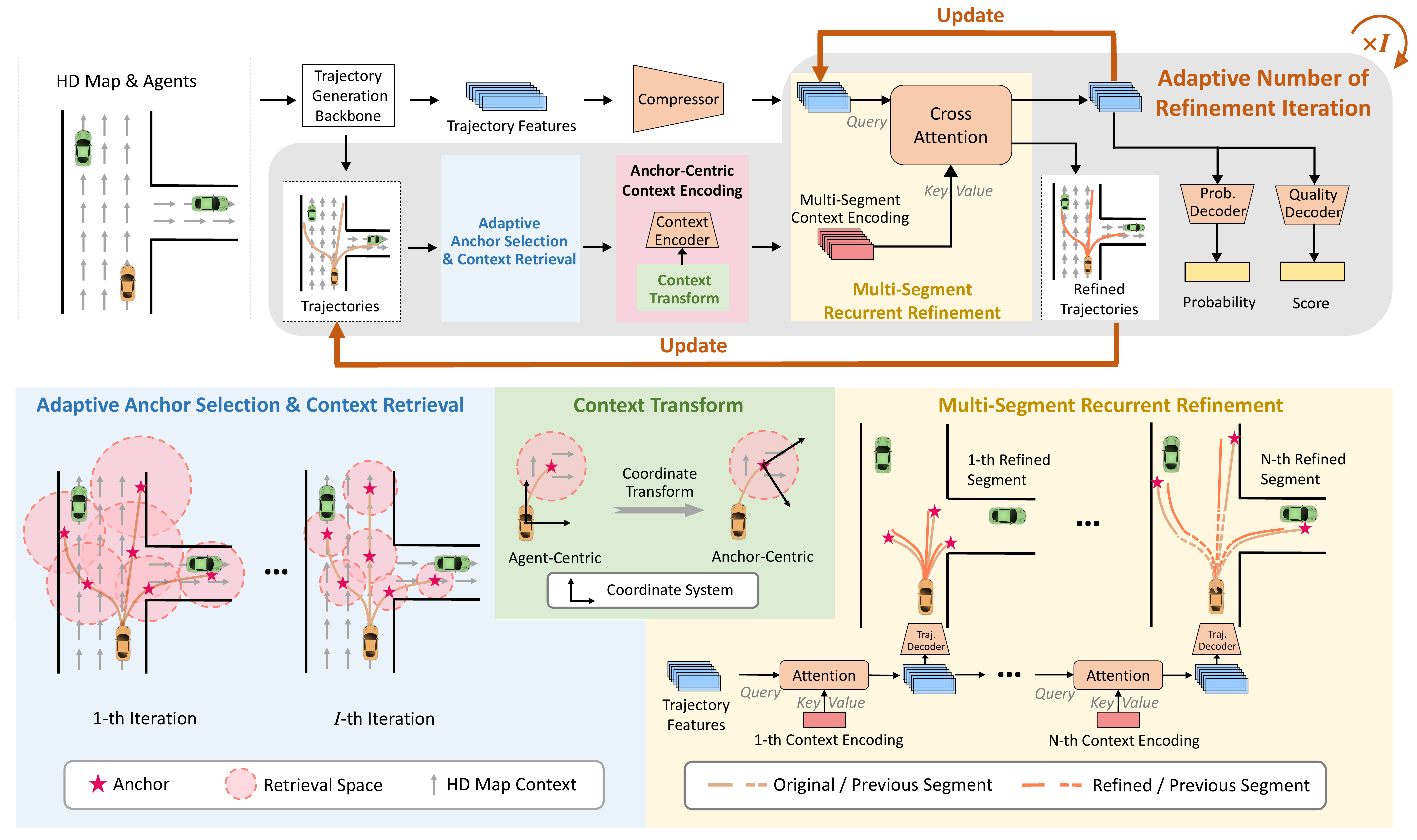}
  \vspace{-5pt}
  \small\caption{
Overview of our framework. The top section concisely illustrates the full pipeline, while the lower section introduces details of three core modules. We first pass HD map and agent information to a prediction model backbone, generating initial trajectories and trajectory features. The initial predicted trajectory is then used to adaptively select anchors and retrieve critical context elements (bottom left). The contexts retrieved by each anchor are transformed to the coordinate frame centered at the corresponding anchors (bottom middle). The encoded contexts are then utilized to refine each trajectory segment by generating the offset of each trajectory segment (bottom right). Our model will predict a quality score measuring the prediction quality, and adaptively decide the number of refinement iterations (upper right). Our method is lightweight and can be seamlessly integrated with most existing motion prediction models.
  }
  \label{fig:pipeline}
  \vspace{-7pt}
\end{figure*}

SmartRefine is a scenario-adaptive refinement method to enhance the performance of motion prediction models with limited additional computation by adaptively iterating between retrieving critical context information and predicting more accurate trajectories. The overall structure is illustrated in Fig.~\ref{fig:pipeline} and Algorithm~\ref{algo:inference}. In Sec.~\ref{sec:problem_formulation}, we introduce the formulation of the problem. In Sec.~\ref{sec:adaptive refine}, we elaborate on the proposed methodologies. In Sec.~\ref{sec:train}, we introduce the training details of our framework.

\subsection{Problem formulation}
\label{sec:problem_formulation}
Given the observed states of the target agent $\mathbf{s}_h=[s_{-T_h}, s_{-T_h+1},\dots,s_0]$ in the historic $T_h$ steps, we aim at predicting its future states $\mathbf{s}_f=[s_{1}, s_{2},\dots,s_{T_f}]$ of $T_f$ steps and associated probabilities $\mathbf{p}$. Naturally, the target agent will interact with the context $\mathbf{c} = (\mathbf{o}_h, \mathbf{m})$, including historic states of surrounding agents $\mathbf{o}_h$, and the HD map $\mathbf{m}$.
For the HD map, we adopt the vectorized representation ~\citep{gao2020vectornet,liang2020learning}, where each lane is defined as a sequence of points along its centerline. Each point is represented by the coordinates and semantic information. Thus typical motion prediction task is formulated as $(\mathbf{s}_f, \mathbf{p})=f(\mathbf{s}_h, \mathbf{c})$, where $f$ denotes the prediction model.

In this work, the introduced refinement strategy will slightly modifies problem formulation. As in Fig.~\ref{fig:pipeline}, we first have a backbone model $f_b$ generating initial trajectories $\mathbf{s}_f^0$ and trajectory features $\mathbf{h}_f^0$, as typical prediction methods $(\mathbf{s}_f^0, \mathbf{h}_f^0, \mathbf{p})=f_b(\mathbf{s}_h, \mathbf{c})$. The initial predicted trajectory $\mathbf{s}_f^0$ is then used to select anchors $\mathbf{a}^0$ and retrieve critical context elements $\mathbf{c}^0$, which are then passed along with the trajectory features $\mathbf{h}_f^0$ into the refinement model $f_r$ for trajectory refinement. Note that the refinement can be multi-iteration, thus refinement of the $i$-th iteration is formulated as $(\delta\mathbf{s}_f^{i}, \mathbf{h}_f^{i}, \mathbf{p})=f_r(\mathbf{h}_f^{i-1}, \mathbf{s}_f^{i-1}, \mathbf{c}^{i-1})$. 
The generated offset $\delta \mathbf{s}_f^{i}$ is added to the input trajectory $\mathbf{s}_f^{i-1}$ for refinement.

\subsection{Scenario-Adaptive Refinement}
\label{sec:adaptive refine}

\subsubsection{Adaptive Anchor/Context Selection}
\label{sec:anchor&radius}
Trajectory refinement necessitates informative anchor selection and context retrieval.
After the backbone generates the initial trajectories and trajectory features, SmartRefine first selects an adaptive number of anchors along the trajectories, and then retrieves contexts near the anchors with adaptive radius (illustrated in the bottom left in Fig.~\ref{fig:pipeline}).

\vspace{1mm}

\noindent \textbf{Anchor Selection.} To select anchors along the trajectory, an intuitive option is to choose the last waypoint of the trajectory as the anchor for long-term context retrieval, as in goal-based predictions~\citep{zhao2021tnt,wang2023ganet}. However, relying exclusively on the endpoint as an anchor often misses the intermediate evolution and progression of the trajectory, resulting in deviations and oscillations. On the other extreme, selecting all $T_f$ waypoints as anchors will lead to large but unnecessary computations. To balance between context richness and computation efficiency, we adopt an adaptive approach, where the trajectory is divided into $N<T_f$ segments and each segment's endpoint is selected as an anchor. 
Note that motion prediction benchmarks typically consider varied prediction horizon lengths.
To depress the cumulative error in long-horizon prediction, the number of segments $N$ is designed to adapt to the prediction horizon. 

\vspace{1mm}

\noindent \textbf{Context Retrieval.} When retrieving context information $\mathbf{c}$ around the anchors, previous methods usually employ a fixed radius or rectangle around the anchor to extract local contexts (e.g. map and nearby agents)~\citep{shi2022motion,choi2022r}. Such fixed retrieval strategies, though straightforward, can be sub-optimal especially when applied across diverse datasets and scenarios which could require different retrieval ranges. 
To address this issue, our idea is that, the retrieval range of one anchor should depend on 1) the refinement iteration $i$, because only fine-grained context information is needed in later refinement iterations, where the trajectories are more accurate compared to early iterations.
2) the target agent's speed $v$ around the anchor, since large speed requires a far and long-term vision of the environment context.
Thus we introduce an adaptive retrieval strategy, where each anchor's retrieval range $R$ varies by the refinement iterations $i$ and the agent's average velocity $v$ around the anchor: $R_{i, v}=\mathcal{F}(i)\cdot{v}$. 
$\mathcal{F}(i)$ can be any monotonically decreasing function of the refinement iteration $i$, to reduce the retrieval range as the trajectory becomes more precise. We instantiate it with an exponential decay function $\mathcal{F}(i)=\beta(\frac{1}{2})^{i-1}$, where $\beta$ denotes a constant. The retrieval range is also constrained within a permissible range $[R_{\min}, R_{\max}]$ to ensure computational stability and efficiency. 

\subsubsection{Anchor-Centric Context Encoding}

\label{sec:encoding&attn}

With the retrieved context elements, most existing methods directly use their embeddings 
generated by the prediction backbone for further refinement~\citep{zhou2023query}. However, these context embeddings from the backbone are typically encoded in the coordinate frame centered at the target agent's current position. While this agent-centric embedding emphasizes the context details around the target agent, the refinement process instead desires the context nuances along the future trajectories. To address this issue, we propose an anchor-centric context encoding approach, to better capture future trajectory details. (illustrated in the bottom middle in Fig.~\ref{fig:pipeline})
This is achieved by transforming the context features into the coordinate frame centered and aligned with the anchor, and then encoding the features to generate anchor-centric context embeddings for refinement.
Note that as the positions and orientations of anchors are dynamically changing across different refinement iterations $i$, these anchor-centric context features could also vary.
This encoding process is applied to every context element retrieved around the anchor $a$, generating a set of context embeddings $\mathbf{z}_a$.

\subsubsection{Recurrent and Multi-Iteration Refinement}
\label{sec:rec}
The anchor-centric context embeddings introduced in the previous section are then fused with the trajectory embeddings to refine predictions. To this end, previous methods usually fuse all embeddings once and refine the trajectory of the whole future horizon, which suffers from long-term cumulative error. To mitigate this issue, 
we adopt a recurrent refinement strategy where we divide the trajectory into $N$ segments, and refine the trajectory $N$ times, one trajectory segment each time (illustrated in the bottom right in Fig.~\ref{fig:pipeline}). Note that here $N$ equals the number of anchors, which means each trajectory segment corresponds to one anchor. For each trajectory segment, we refine it only using the context retrieved by the corresponding anchor to enhance local context fusion. Besides, refining all $N$ segments finishes one iteration of refinement, and we will conduct multiple refinement iterations.

\vspace{0.5mm}
\noindent \textbf{Recurrent Refinement.} Specifically, in each recurrent refinement step, we refine one trajectory segment, with a set of scene context embeddings $\mathbf{z}_a$ around the segment's corresponding anchor $a$, and the target agent's future trajectory embeddings $\mathbf{h}_f$.
We adopt the cross attention mechanism~\citep{vaswani2017attention} to fuse them, where the trajectory embeddings are used as queries to attend to the keys/values from the context embeddings $\mathbf{z}_a$. The fused trajectory embeddings will be used to predict $\delta\mathbf{s}$, the offset of waypoints in the trajectory segment, to subtly adjust the original trajectory segment. The updated trajectory embeddings will also be leveraged as new queries to refine the next segment. 
After all $N$ steps within one iteration, the whole trajectory will be adjusted, and the trajectory embeddings will also be updated with rich context information. 
Note that we predict multiple possible predictions, thus the trajectory embeddings will also be used to predict the probability of each prediction.

\vspace{0.5mm}
\noindent  \textbf{Multi-Iteration Refinement.}
After one iteration terminates, the updated trajectory and trajectory embeddings will be used to start another refinement iteration. We first repeat the adaptive anchor selection and context retrieval mentioned in Sec.~\ref{sec:anchor&radius}, and then conduct another $N$ recurrent refinement steps. 
Thus in the multi-iteration refinement, the trajectory and trajectory embeddings used in one iteration can come from either the backbone (for the first iteration) or the previous refinement iteration (for the later iterations). Note that in the first refinement iteration, we employ a compressor network to reduce the hidden dimension of the trajectory embedding from the backbone for efficient refinement, which we found brings little performance loss.

\begin{algorithm}[t]
\caption{Adaptive Inference with SmartRefine}
\label{algo:inference}
    \begin{algorithmic}[1]
        \Require
        Backbone model $f_b$, refinement model $f_r$, quality score decoder $f_d$, agents history trajectories $\mathbf{s}_h$,\ scene context $\mathbf{c}$, and score threshold $\Bar{q}$,\ maximum refinement iteration at inference $I'$. 
        \Ensure
        Target agent's future trajectories $\mathbf{s}_f$ and corresponding probabilities $\mathbf{p}$.
        \State $\mathbf{s}_f^0, \mathbf{h}_f^0, \mathbf{p}^0 = f_b(\mathbf{s}_f^0, \mathbf{c})$
        \rightcomment{Backbone prediction}
        \State $q^0 = f_d(\mathbf{h}_f^0)$
        \rightcomment{Initial quality score}
        \If{$\color{blue} q^0 > \Bar{q}$} \rightcommentblue{No need for refinement}
            \State \textbf{Return} $\mathbf{s}_f^0, \mathbf{p}^0$
        \EndIf
        \For{$i=1,2,\dots,I'$}\rightcomment{Multi-iteration refine}
            \State Adaptively select anchors and contexts
            \State Adpatively encode contexts $\mathbf{c}^{i-1}$ (anchor-centric)
            \State Multi-segment recurrent refinement:
                \markcomment{3}{$\delta \mathbf{s}_f^i, \mathbf{h}_f^i, \mathbf{p}^i, q^i = f_r(\mathbf{s}_f^{i-1}, \mathbf{h}_f^{i-1}, 
                \mathbf{c}^{i-1})$}
                \markcomment{3}{$\mathbf{s}_f^i = \mathbf{s}_f^{i-1} + \delta \mathbf{s}_f^i $}
            \If{$\color{blue} q^i < q^{i-1}$}    \rightcommentblue{Terminate refinement}
                \State \textbf{Return} $\mathbf{s}_f^i, \mathbf{p}^i$
            \EndIf
        \EndFor
        \State \textbf{Return} $\mathbf{s}_f^i, \mathbf{p}^i$
        \rightcommentblue{Reach max refinement iteration}
    \end{algorithmic}
\end{algorithm}

\subsubsection{Adaptive Number of Refinement Iterations}
\label{sec:refine}
As mentioned in the previous section, the refinement is multi-iteration. As we get into later iterations, the predicted trajectory becomes more accurate.
Nonetheless, there's a diminishing return as we utilize more refinement iterations.  To trade-off between performance gain and additional computation, we propose an adaptive refinement strategy where the number of refinement iterations is dynamically adapted according to the current prediction quality and remaining potential refinement improvement. 
Specifically, we propose a quality score to quantify current prediction quality.

\vspace{1mm}
\noindent \textbf{Quality Score Design.} At the training stage, since we have access to the ground-truth trajectory and predicted trajectory of all refinement iterations, an intuitive measure for the quality of the predicted trajectory in iteration $i$ is $d_{max} - d_i$, where $d_{max}$ denotes the largest predicted error among all iterations. A small $d_i$ means that the current prediction has already been improved from the largest prediction error $d_{max}$, and thus of high quality. 
However, this design as the quality score can be unstable as it lies in a big range and can vary a lot in different scenarios.
We then normalize $d_{max} - d_i$ with $d_{max}-d_{min}$, where $d_{min}$ denotes the smallest predicted error among all iterations.
Thus the final quality score lies in between [0,1] and is designed as
\begin{equation}
    q_i=\frac{d_{max} - d_i}{d_{max}-d_{min}}
    \label{eq:score}
\end{equation}

To enable the model to predict the quality score, we first employ a GRU~\citep{cho2014learning} to recurrently process the trajectory embeddings of all previous iterations, and then use an MLP to generate the quality score. 
Thus our model will generate three types of output: the predicted trajectories, the probabilities of the trajectories, and the quality score. 

\vspace{1mm}
\noindent \textbf{Adaptive Refinement Iteration.}
The contents above introduce how to design and generate the quality score at the training stage. At the inference stage, the initial trajectory features from the backbone will be appended with the quality score decoder to predict an initial quality score. Our model will also predict the quality score at each refinement iteration. Thus we propose a simple yet effective strategy to dynamically decide whether another refinement iteration is needed.
As outlined in Algorithm~\ref{algo:inference}, the adaptive strategy includes three major criteria: 1) an initial quality score threshold is set, and we only conduct refinement if the initial predicted quality score is below the threshold, which means the current prediction is not good enough and need refinement; 2) at any of the later refinement iterations, if the quality score stops growing compared to previous iterations, we terminate the refinement as this indicates that we are close to ideal performance, and further refinement will only bring diminishing returns or negative effects due to over-refinement; 3) a maximum refinement iteration is set, which is a hyper-parameter to balance performance and efficiency.

\subsection{Training Loss}
\label{sec:train}
The training of our model considers three loss terms, and uses hyper-parameter $\alpha$ to balance them:
\begin{equation}
    \mathcal{L} = \mathcal{L}_{\text{cls}} + \mathcal{L}_{\text{reg}} + \alpha\cdot\mathcal{L}_{\text{score}}
    \label{eq:loss}
\end{equation}
where $\mathcal{L}_{\text{cls}}$ denotes the cross-entropy classification loss for predicting the probability of the multi-modal trajectories. 
For the regression loss $\mathcal{L}_{\text{reg}}$, as our model predicts a Laplace distribution for each time step's waypoint, we calculate the negative log-likelihood of the ground-truth trajectory in the predicted distribution. Note that among all predicted multi-modal trajectories, only the modal closest to the ground truth is considered for the regression loss.
For the quality score loss $\mathcal{L}_{\text{score}}$, in the training stage, we fix the refinement iteration as $I$ and each iteration will output a quality score. Thus for each iteration $i$, we calculate the $\ell_1$ loss between the predicted quality score $\hat{q}_i$ and labeled quality score $q_i$ (see Sec~\ref{sec:refine}), and average the loss over all iterations:
\begin{equation}
\mathcal{L}_{\text{score}} = \frac{1}{I+1}\sum_{i=0}^{I}\| \hat{q}_i - q_i\|_1
    \label{eq:score_loss}
\end{equation}
Similarly, the score loss also only considers the modal closest to the ground truth.

\section{Experiments}

\begin{table*}[!t]

\small
\def\arraystretch{1.05} 
    \vspace{-3mm}
    \centering
    \small
    \resizebox{0.75\linewidth}{!}{
    \begin{tabular}{cccccccc}
        \toprule
            Dataset & Method  & minFDE $\downarrow$     & minADE $\downarrow$  & MR $\downarrow$  & \#Param.(M) $\downarrow$ & Flops(G) $\downarrow$ & Latency(ms) $\downarrow$    \\ 
            \midrule
            \multirow{7}{*}{Argoverse} & HiVT~\citep{zhou2022hivt}
            & 0.969 & 0.661  & 0.092 & 2.5 & 2.6 & 54$\pm$4.0 \\
            & \cellcolor{lightgrey}HiVT w/ Ours
            
            & \cellcolor{lightgrey}0.911 & \cellcolor{lightgrey}0.646 & \cellcolor{lightgrey}0.083 & \cellcolor{lightgrey}2.7 & \cellcolor{lightgrey}2.7 & \cellcolor{lightgrey}67$\pm$8.4 \\ 
            \cmidrule(r){2-2}
             & Prophnet*~\citep{wang2023prophnet}
            & 1.004  &  0.687   & 0.093 & 15.2&7.8&59$\pm$1.7 \\ 
            & \cellcolor{lightgrey}Prophnet w/ Ours
            
            & \cellcolor{lightgrey}0.967 & \cellcolor{lightgrey}0.675  & \cellcolor{lightgrey}0.092 & \cellcolor{lightgrey}15.4
            &\cellcolor{lightgrey}7.9 & \cellcolor{lightgrey}71$\pm$6.2 \\
            \cmidrule(r){2-2}
            & mmTransformer~\citep{liu2021multimodal}
            & 1.081  &  0.709   & 0.102 & 2.6 & 1.2& 15$\pm$4.8\\ 
             & \cellcolor{lightgrey}mmTransformer w/ Ours
             
            & \cellcolor{lightgrey}1.023 &\cellcolor{lightgrey} 0.692  & \cellcolor{lightgrey}0.094 & \cellcolor{lightgrey}2.8& \cellcolor{lightgrey}1.3& \cellcolor{lightgrey}27$\pm$9.7\\

        \midrule
        \multirow{7}{*}{Argoverse 2} & DenseTNT~\citep{gu2021densetnt} 
        & 1.624      & 0.964 & 0.233 & 1.6& 3.6
        &1{,}075$\pm$199 \\
           & \cellcolor{lightgrey}DenseTNT w/ Ours
           
           & \cellcolor{lightgrey}1.553  & \cellcolor{lightgrey}0.834  & \cellcolor{lightgrey}0.221 &  \cellcolor{lightgrey}1.9& \cellcolor{lightgrey}4.0 & \cellcolor{lightgrey}1{,}099$\pm$212 \\ 
           \cmidrule(r){2-2}   
            & QCNet (no ref)   
            & 1.304   &  0.729           & 0.164 & 5.5 & 47.0 & 338$\pm$53
            \\ 
           & \cellcolor{lightgrey}QCNet (no ref) w/ Ours
            
           & \cellcolor{lightgrey}1.258     & \cellcolor{lightgrey}0.718  & \cellcolor{lightgrey}0.157 & \cellcolor{lightgrey}5.8 & \cellcolor{lightgrey}47.4 &  \cellcolor{lightgrey}363$\pm$67 \\ 
            \cmidrule(r){2-2} 
           & QCNet~\citep{zhou2023query} & 1.253      & 0.720                & 0.157 & 7.7 &55.8&392$\pm$54\\ 
           & \cellcolor{lightgrey}QCNet w/ Ours       
           
           & \cellcolor{lightgrey}1.240  & \cellcolor{lightgrey}0.716  & \cellcolor{lightgrey}0.156 & \cellcolor{lightgrey}8.0 & \cellcolor{lightgrey}56.2 & 
           \cellcolor{lightgrey}418$\pm$68 \\  
        \bottomrule
    \end{tabular}
    }
    \vspace{-0.75em}
    \caption{Performance and computation efficiency on Argoverse and Argoverse 2 val set. Methods that are not open-source and reproduced by us are marked with the symbol ``*''. QCNet (no ref) denotes the version where we remove the original refinement module in the QCNet.
    SmartRefine consistently improves the accuracy of all considered state-of-the-art methods with limited added computation and latency.
    }
    \label{tab:resultself}
\end{table*}

We first report the prediction accuracy and computation cost on the validation set of the two datasets. As shown in Table~\ref{tab:resultself}, SmartRefine can consistently improve the prediction accuracy of all considered state-of-the-art methods with limited added parameters, Flops, and latency.

\subsection{Experimental Settings}
\noindent \textbf{Dataset.} We train and evaluate our method on two large-scale motion forecasting datasets: Argoverse~\citep{chang2019argoverse} and Argoverse 2~\citep{wilson2023argoverse}. Argoverse contains 333k scenarios collected from interactive and dense traffic. Each scenario provides the HD map and 2 seconds of history trajectory data, to predict the trajectory for future 3 seconds, sampled at 10Hz. Following the official guide, we split the training, validation, and test set to 205k, 39k, 78k scenarios respectively. Argoverse 2 upgrades the previous dataset to include 250K sequences with higher prediction complexity. It extends the historic and prediction horizon to 5 seconds and 6 seconds respectively, sampled at 10Hz. The data is split into 200k, 25k, and 25k for training, validation, and test respectively.
\vspace{1mm}

\noindent \textbf{Metrics.} Following the official dataset settings, we evaluate our model on the standard metrics for motion prediction, including minimum Average Displacement Error (minADE), minimum Final Displacement Error (minFDE), and Miss Rate (MR). These metrics allow models to forecast up to 6 trajectories for each agent and define the one that has the minimum endpoint error as the best trajectory to calculate the above metrics.
\vspace{1mm}
 
\noindent \textbf{Baselines.} As mentioned earlier, our SmartRefine can be seamlessly integrated into most existing trajectory prediction methods to improve accuracy with limited additional computation. In our experiment, we consider five popular and state-of-the-art methods as the prediction backbone to evaluate how our SmartRefine further improves the performance: HiVT~\citep{zhou2022hivt}, Prophnet~\citep{wang2023prophnet}, mmTransformer~\citep{liu2021multimodal}, DenseTNT~\citep{gu2021densetnt} and QCNet~\citep{zhou2023query}. 
Note that, since QCNet includes a refinement module itself, we introduce another baseline 
QCNet (no refinement) which removes the original refinement module from the QCNet. 
For implementation and computation details of our methods, please refer to the supplementary material.

\subsection{Quantitative Result}
\label{sec:quantitative_result}

\noindent \textbf{Performance on Val Set.} We first report the prediction accuracy and computation cost on the validation set of the two datasets. As shown in Table~\ref{tab:resultself}, SmartRefine can consistently improve the prediction accuracy of all considered state-of-the-art methods with limited added parameters, Flops, and latency. For instance, our method can reduce the minFDE of Prophnet, mmTransformer, HiVT, DenseTNT, and QCNet (no ref) by 6.0\%, 3.7\%, 5.4\%, 4.3\%, 3.5\% respectively.
Besides, our refinement model only added Flops by 130M on Argoverse and 408M on Argoverse 2, and thus is much smaller than the Flops of the backbone which range from 1.2G to 55.8G. 
Our method benefits QCNet less than other methods, because QCNet itself incorporates a relatively large refinement network that has 2,200K parameters. However, if we replace the 
refinement module in QCNet with ours, we can achieve competitive results with much less added parameters, flops, and latency.
\vspace{1mm}

\noindent \textbf{Performance on Test Set.}
We also submit our method to Argoverse and Argoverse 2 test set and the results are shown in Table~\ref{tab:resultothers}.
Again, SmartRefine consistently improves the prediction accuracy of all considered methods. For instance, SmartRefine reduces the minFDE of HiVT, Prophnet, mmTransformer, DenseTNT, and QCNet (no ref) by 3.4\%, 6.9\%, 7.5\%, 4.2\%, 3.9\% respectively. 
Specifically, our SmartRefine based on QCNet outperforms all published ensemble-free works on the Argoverse 2 leaderboard (single agent track) at the time of the paper submission. 
See Sec.~\ref{sec:lead} for a more detailed explanation.

\subsection{Ablation studies}
\label{sec:abi_study}
We conducted comprehensive ablation studies on the effect of each component in our proposed method, on the Argoverse validation set. We show the performance when we add SmartRefine on top of HiVT, and the results based on other backbones can be found in the supplementary material.
\vspace{1mm}

\begin{figure}[!t]
    \centering
    \includegraphics[width=\linewidth]{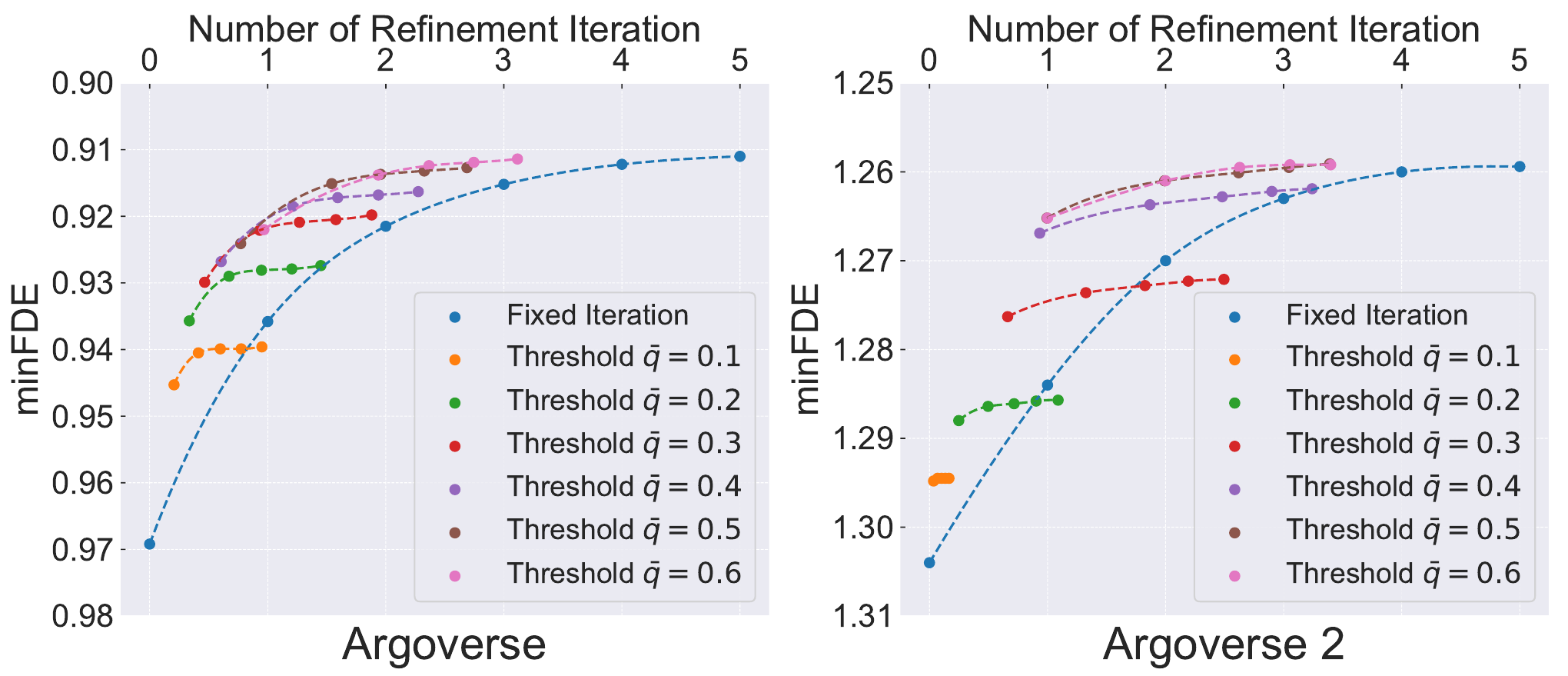}
    \vspace{-2em}
    \caption{Comparison between the fixed and adaptive number of refinement iterations. For the adaptive methods, We tested different quality score thresholds $\Bar{q}$ mentioned in Algorithm~\ref{algo:inference}.}
    \label{fig:adaptive}
    \vspace{-1em}
\end{figure}

\noindent\textbf{Adaptive Refinement Iterations.} Fig.~\ref{fig:adaptive} ablates the effect of adaptive refinement iterations. 
When we use a fixed number of iterations (blue curve), the prediction accuracy improves with more refinement iterations, but diminishing or zero improvements are witnessed in later iterations. In comparison, when we adopt the adaptive refinement iteration proposed by SmartRefine, we can achieve higher accuracy with a smaller number of refinement iterations. Note that we studied different quality score thresholds $\Bar{q}$ for adaptive refinement, while a higher threshold leads to better accuracy, we find $\Bar{q}=0.5$ achieves an ideal performance and higher thresholds perform similarly. Besides, for each threshold, we ablate different limits for maximum refinement iteration, resulting in 5 points for each curve.
\vspace{1mm}

\noindent\textbf{Anchor Numbers.} Table~\ref{tab:anchor} ablates the effect of anchor numbers. We can see that increasing the anchor number from 1 to 2 reduces the minFDE. However, excessively increasing the number of anchors is ineffective, as it brings much larger model parameters with the same or slightly worse accuracy.

\vspace{1mm}

\begin{table}
\small
\vspace{-3mm}
    \centering
    \small
    \resizebox{1.0\linewidth}{!}{
    \begin{tabular}{ccccc}
        \toprule                         
            Dataset & Method& minFDE $\downarrow$    & minADE $\downarrow$  & MR  $\downarrow$    \\
            \midrule
            \multirow{6}{*}{Argoverse} & HiVT~\citep{zhou2022hivt}      & 1.17         & 0.77          & 0.13 \\ 
            & \cellcolor{lightgrey}HiVT w/ Ours 
             
            & \cellcolor{lightgrey}1.13 & \cellcolor{lightgrey}0.77 & \cellcolor{lightgrey}0.12  \\ 
            \cmidrule(r){2-2}
           & ProphNet*~\citep{wang2023prophnet}     & 1.30 & 0.85& 0.14\\
            & \cellcolor{lightgrey}Prophnet* w/ Ours 
             
            & \cellcolor{lightgrey}1.21 & \cellcolor{lightgrey}0.81 & \cellcolor{lightgrey}0.13\\ 
            \cmidrule(r){2-2}
            &
           mmTransformer~\citep{liu2021multimodal}
            & 1.34          & 0.84          & 0.15          \\
             & \cellcolor{lightgrey}mmTransformer w/ Ours
              
             & \cellcolor{lightgrey}1.24& \cellcolor{lightgrey}0.81& \cellcolor{lightgrey}0.14 \\
           
           \midrule
           \multirow{6}{*}{Argoverse 2}&
           DenseTNT~\citep{gu2021densetnt}
            & 1.66          & 0.99          & 0.23          \\
            & \cellcolor{lightgrey}DenseTNT w/ Ours
            
            & \cellcolor{lightgrey}1.59& \cellcolor{lightgrey}0.85&\cellcolor{lightgrey}0.22 \\
            \cmidrule(r){2-2}
            
           & QCNet (no ref) 
           & 1.29  & 0.65 & 0.16 \\
           & \cellcolor{lightgrey}QCNet (no ref) w/ Ours 
           
           & \cellcolor{lightgrey}1.24 & \cellcolor{lightgrey}0.64 & \cellcolor{lightgrey}0.15\\ 
            \cmidrule(r){2-2}
            
            & QCNet~\citep{zhou2023query}      &   1.24       & 0.64        &0.15 \\ 
           & \cellcolor{lightgrey}QCNet w/ Ours  
           
           &\cellcolor{lightgrey}1.23 
           & \cellcolor{lightgrey}0.63& \cellcolor{lightgrey}0.15 \\ 
        \bottomrule
    \end{tabular}
    }
    \vspace{-0.75em}
    \caption{Performance on Argoverse (1\&2) test set. Methods that are not open-source and reproduced by us are marked with the symbol ``*''. The original DenseTNT and QCNet (no ref) did not report results in the test set, thus we train them to match the validation set and submit them to the test set.
    }
\label{tab:resultothers}
\end{table}

\begin{figure*}[!h]
    \centering
    
    \includegraphics[width=0.9\textwidth]{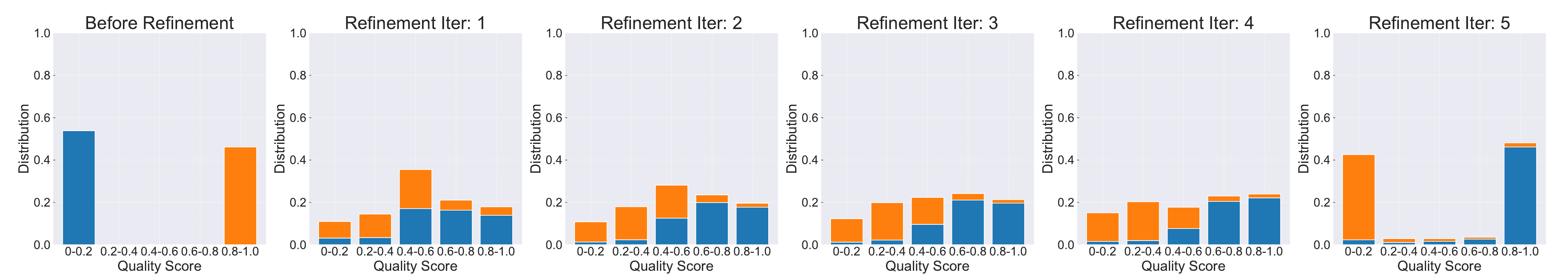}
    \vspace{-3mm}
    \caption{A study to understand the mechanism behind the refinement. 
    Specifically, We mark the quality score distribution of the predictive trajectory before refinement, and track how the quality score changes along the multi-iteration refinement. We can see while the overall performance is improved, not every trajectory benefits from refinement, which implies the necessity of adaptive refinement. See Sec.~\ref{sec:discussion} for detailed discussions.
    }
    \label{fig:score}
    \vspace{-1.5em}
\end{figure*}

\noindent\textbf{Context Representation.} In Table~\ref{tab:context}, we compare the performance when we employ the fixed agent-centric context embeddings or the adaptive anchor-centric context embeddings (see details in Sec~\ref{sec:encoding&attn}). We can see our adaptive anchor-centric encoding effectively outperforms the fixed agent-centric context encoding.
\vspace{1mm}

\noindent\textbf{Retrieval Radius.} Table~\ref{tab:radius} studies the effect of the radius used to retrieve contexts. We can see a fixed retrieval radius from 50 to 2 can be sub-optimal, as a large retrieval radius might lead to redundant or unrelavant context information, while a small radius might not provide sufficient context for refinement.
In comparison, our SmartRefine adapts the radius to each agent's velocity, and decays the radius with the number of refinement iterations (see details in Sec~\ref{sec:anchor&radius}). Here we consider two strategies for radius decay: linear decay and exponential decay. Both methods lead to lower minFDE compared to the fixed radius, and the exponential decay has lower Flops compared to linear decay.
\vspace{1mm}

\subsection{Discussion}
\label{sec:discussion}

While many previous works have explored various refinement strategies, few of them explain how exactly the refinement works and what limitations it has. For example, 
it remains unclear whether every trajectory is improved by the refinement. In this section, we take a step forward to study the mechanism behind the refinement. Specifically, for each predicted trajectory, we measure its accuracy using the quality score proposed earlier.
We will track how the quality score changes along the multi-iteration refinements.
As shown in Fig.~\ref{fig:score}, before the refinement, the quality score of the trajectories most fall in two categories: about 56\% of the trajectories have a low quality score between 0 and 0.2 (marked as blue), and about 44\% trajectories have a high quality score between 0.8 and 1 (marked as orange). We then conduct multi-iteration refinement and track how the quality score of the "blue" and "orange" trajectories change. Two phenomenons are observed:

\begin{itemize}
    \item \textit{Not every trajectory benefits from refinement}: as the refinement goes, we can see that the initially low-score "blue" trajectories gradually move to the right, which means they become more accurate. However, the initially high-score "orange" trajectories gradually move to the left, which means they become less accurate. In fact, we can see that initially low-score blue trajectories and initially high-score orange trajectories switched their quality score at the final iteration.
    We hypothesize that this is because while inaccurate trajectories can get much improvement from refinement, the accurate trajectories are good enough and can hardly be further improved, 
    or even be pushed away from the ground truth due to over-refinement.
    \item \textit{The overall performance is better:} though the initially accurate and inaccurate trajectories switch their quality scores at the final iteration, we see the final iteration has a higher percentage of high-score trajectories. This is why our refinement can lead to a higher overall performance.
    \end{itemize}
These results show the necessity of adaptive refinement.

\begin{table}[!t]
\begin{minipage}{0.48\linewidth}
    \centering
        \resizebox{\linewidth}{!}{
    \begin{tabular}{c|cc}
        \toprule
            \multicolumn{1}{c|}{\#Anchor numbers}   & minFDE & \#Param.      \\ \midrule
         1 & 0.928 & 134K \\
         2 & 0.911 & 207K\\
         3 & 0.911 & 280K\\
         5 & 0.915 & 433K \\
         6 & 0.916 & 509K  \\
         \bottomrule
    \end{tabular}
    }
    \vspace{-1em}
    \small\caption{Ablation study on the number of anchors.}
    \vspace{-1em}
    \label{tab:anchor}
\end{minipage}\hfill
\begin{minipage}{0.48\linewidth}
        \centering
    \resizebox{\linewidth}{!}{
    \begin{tabular}{c|c}
        \toprule
          Context Encoding  & minFDE      \\ \midrule
         Agent-Centric & 0.941  \\
         Anchor-Centric & 0.911 \\
         \bottomrule
    \end{tabular}
    }
    \vspace{-1em}
    \small\caption{Ablation study on how the contexts are encoded.}
    \label{tab:context}
\end{minipage}
\end{table}

\begin{table}[t!]
    \centering
    \resizebox{0.8\linewidth}{!}{
    \begin{tabular}{c|c|c|c}
          \toprule
          
          & Retrieval Radius   & minFDE &Flops (M)      \\ \midrule
         \multirow{4}{*}{\tabincell{c}{Fixed Radius}} 
         &50 & 0.926&2,297  \\
         &20 & 0.923& 722 \\
         &10 & 0.921&325 \\
         &2 & 0.930&58 \\
         \midrule
         \multirow{2}{*}{\tabincell{c}{Adaptive Radius}} 
         &$R_{max}$=10, $R_{min}$=2, linear & 0.911&245 \\
         &$R_{max}$=10, $R_{min}$=2, exp~~~ & 0.911&130\\
         \bottomrule
    \end{tabular}
    }
    \vspace{-0.8em}
    \caption{Ablation study on the context retrieval radius. Linear and exp denote different ways to decay the radius.}
    \label{tab:radius}
\end{table}

\section{Conclusion}
In this paper, we introduce SmartRefine, a scenario-adaptive refinement framework for efficient refinement of motion prediction models. Our method adopts adaptive strategies anchor selection, retrieval radius, and context encoding to conduct refinement that better suits each scenario.
We then propose the quality score to indicate current prediction quality and potential refinement improvement, and adaptively decide how many refinement iterations are needed for each scenario.
Our SmartRefine is not only lightweight, but can also be seamlessly plugged into most existing motion prediction models as it is decoupled from the prediction backbone and only relies on a generic interface between the backbone.
Extensive experiments demonstrate the effectiveness of our approach in terms of both prediction accuracy and computation efficiency.
We also study the mechanism behind multi-iteration refinement.
In the future, we will further extend our framework to multi-agent joint prediction settings. 

{
    \small
    \bibliographystyle{ieeenat_fullname}
    \bibliography{main}
}

\clearpage
\appendix
\maketitlesupplementary

The overall structure of the supplementary material is listed as follows:

$\triangleright$ Sec.~\ref{sec:impl}: \textit{Implementation details of our model.}

$\triangleright$ Sec.~\ref{sec:eva}: \textit{Metric details for evaluation.}

$\triangleright$ Sec.~\ref{sec:lead}: \textit{Performance on Argoverse 2 leaderboard.}

$\triangleright$ Sec.~\ref{sec:abi_ref}: \textit{Comparison with other refinement methods.}

$\triangleright$ Sec.~\ref{sec:abi_all}: \textit{Ablation studies of all backbones.}

$\triangleright$ Sec.~\ref{sec:score_all}: \textit{The quality score distribution of all methods.}

$\triangleright$ Sec.~\ref{sec:vis}: \textit{Visualization results of refinement.}

\section{Implementation Details}
\label{sec:impl}

\subsection{Backbones}
We apply our SmartRefine to six classic and state-of-the-art motion prediction backbones: 
HiVT~\citep{zhou2022hivt}, Prophnet~\citep{wang2023prophnet}, mmTransformer~\citep{liu2021multimodal}, DenseTNT~\citep{gu2021densetnt}, QCNet~\citep{zhou2023query}, and QCNet (no refine)~\citep{zhou2023query}. 
For implementation, we reproduce Prophnet since it is not open-source, and utilize the open-source codes for all other backbones. 
In the rest of this section, we introduce the implementation details of our SmartRefine.

\subsection{Compressor}
We use an MLP network to reduce the hidden dimension of the trajectory embeddings from backbones. The following hyperparameters are used:
\begin{itemize}[leftmargin=2em]
    \item Number of layers: 2
    \item Input size: the original hidden size set by the trajectory generation backbone
    \item Output size: 64
\end{itemize}

\subsection{Context Retrieve and Encoding}
During refinement, we first select anchors along the trajectories, which are used to retrieve contexts. We represent each anchor's retrieval range as $R_{i, v}=\mathcal{F}(i)\cdot{v}$, as mentioned in Sec.~\ref{sec:anchor&radius}. In practice: 1) $\mathcal{F}(i)=\beta(\frac{1}{2})^{i-1}$, and we set radius decay constant $\beta$ as 0.8; 2) the average speed $v$ around one anchor is calculated based on the speed when the agent passes through the anchor's corresponding trajectory segment.

When encoding the context, the early fusion strategy~\cite{nayakanti2023wayformer} is utilized. Specifically, for each context, we first use MLP layers to encode the context components (positions, semantic information, and distance to the anchor) separately, then add them together, and then apply MLP layers to encode the added embedding. We use the following model hyperparameters: 
\begin{itemize}[leftmargin=2em]
    \item Number of Anchors: 2 for Argoverse and 4 for Argoverse 2
    \item Number of encoder layers: 2
    \item Input size of each context: 2 for (x,y) positions and 1 for semantic information
    \item Hidden size: 64
    \item Fusion operator: ``add"
    \item Number of fusion layers: 2
\end{itemize}

\subsection{Cross Attention for Refinement}
We utilize a multi-head attention module to refine each trajectory segment.
It takes the trajectory embeddings as queries and the context encodings as keys/values. We use the following model hyperparameters:

\begin{itemize}[leftmargin=2em]
    \item Number of attention layers: 1
    \item Hidden size: 64
    \item Number of attention heads: 8
    \item Dropout: 0.1
    \item Activation: ReLU
\end{itemize}

\begin{table}[t!]
\small
\vspace{-3mm}
    \centering
    \small
    \resizebox{1.0\linewidth}{!}{
    \begin{tabular}{ccccc}
        \toprule                        
            Rank & Method& minFDE $\downarrow$     & minADE $\downarrow$  & MR  $\downarrow$    \\
            \midrule
            1 & SEPT-iDLab (SEPT)* & 1.15 & 0.61 & 0.14 \\
            2 & GACRND-XLAB (XPredFormer)* & 1.20 & 0.62 & 0.15 \\
            3 & QCNet-AV2 (QCNet) & 1.19 & 0.62 & 0.14 \\
            4 & MTC (MTC)* & 1.17 & 0.61 & 0.14 \\
            5 & Mingkun Wang & 1.19 & 0.62 & 0.14 \\
            6 & AnonNet (AnonNet)* & 1.23 & 0.63 & 0.15 \\
            7 & ls (TraceBack)* & 1.20 & 0.64 & 0.14 \\
            
            \cellcolor{lightgrey}8 & \cellcolor{lightgrey}SmartRefine (ours) & \cellcolor{lightgrey}1.23 & \cellcolor{lightgrey}0.63 & \cellcolor{lightgrey}0.15 \\
        \midrule
        - & QCNet (no ensemble) & 1.24 & 0.64 & 0.15 \\
        - & GANet (published version) & 1.35 & 0.73 & 0.17 \\
        \bottomrule
    \end{tabular}
    }
    \vspace{-1em}
    \caption{Argoverse 2 leaderboard (single agent track) at the time of the paper submission. 
    Unpublished works are marked with the symbol ``*''. 
    Our SmartRefine with QCNet as trajectory generation backbone ranked \#8 on the leaderboard.
    Methods before ours are all unpublished except two methods: 1) The \#3 method is the ensemble version of QCNet, while our method does not utilize ensemble. For a fair comparison, our method outperforms the ensemble-free version of QCNet (as marked in the second last row). 2) The \#5 method is linked to GANet~\citep{wang2023ganet}, which was published before.
    However, the performance in the original published GANet paper (as marked in the last row) is much poorer than that of the \#5 method.
    We suppose the current \#5 performance is obtained by the ensemble version of GANet or an unpublished extended version of GANet.
    Thus our method outperforms all published ensemble-free works on the Argoverse 2 leaderboard (single agent track) at the time of the paper submission.
    }
\label{tab:lead}
\vspace{-1em}
\end{table}

\begin{table*}[t!]

\vspace{-3mm}
    \centering

    \resizebox{0.9\linewidth}{!}{
    \begin{tabular}{cc|cccccc}
        \toprule
            \multicolumn{1}{c}{\multirow{2}{*}{\tabincell{c}{Backbones\\Datasets}}} &
            \multicolumn{1}{c|}{\multirow{2}{*}{Metrics}} &
            \multicolumn{6}{c}{\multirow{1}{*}{\tabincell{c}{Different Ideologies of Refinement Techniques}}} \\
            \cmidrule(r){3-8}
            & & no ref& DCMS$^1$ & QCNet$^1$ & R-Pred$^1$ & MTR$^M$ & Ours$^A$ \\
            \midrule
            \multicolumn{1}{c}{\multirow{2}{*}{\tabincell{c}{HiVT\\Argo}}} &
            minFDE & 0.969 & 0.958& 0.933& 0.929& 0.915&0.911 \\
            & Latency & 54{$\pm$4.0} & 55$\pm$4.4& 64$\pm$5.1& 62$\pm$5.9& 92$\pm$9.4& 67$\pm$8.4 \\
            \midrule
            \multicolumn{1}{c}{\multirow{2}{*}{\tabincell{c}{Prophnet\\Argo}}} &
            minFDE & 1.004 & 0.996 & 0.984& 0.981& 0.968&0.967 \\
            & Latency & 59$\pm$1.7 & 60$\pm$2.4& 68$\pm$3.2& 65$\pm$3.1& 88$\pm$5.9& 71$\pm$6.2 \\
            \midrule
            \multicolumn{1}{c}{\multirow{2}{*}{\tabincell{c}{mmTransformer\\Argo}}} &
            minFDE & 1.081 & 1.066& 1.048& 1.045& 1.022&1.023 \\
            & Latency & 15$\pm$4.8 & 16$\pm$5.5& 22$\pm$5.6& 21$\pm$5.5& 51$\pm$8.4& 27$\pm$9.7 \\
            \midrule
            \multicolumn{1}{c}{\multirow{2}{*}{\tabincell{c}{DenseTNT\\Argo 2}}} &
            minFDE & 1.624 & 1.601& 1.563& 1.576& 1.553&1.553 \\
            & Latency & 1,075$\pm$199 & 1,076$\pm$199& 1,133$\pm$217& 1,085$\pm$209& 1,125$\pm$213& 1,099$\pm$212 \\
            \midrule
            \multicolumn{1}{c}{\multirow{2}{*}{\tabincell{c}{QCNet (no ref)\\Argo 2}}} &
            minFDE & 1.304 & 1.293& 1.253& 1.274& 1.256&1.258 \\
            & Latency & 338$\pm$53 & 339$\pm$53& 392$\pm$54& 348$\pm$55& 387$\pm$62& 363$\pm$67 \\
        \bottomrule
    \end{tabular}
    }

    \caption{
    Comparison of refinement methods. 1/M/A denotes one-iteration, multi-iteration, and adaptive-iteration refinement methods respectively. Our method which utilizes adaptive refinement achieves the best trade-off in minFDE and latency.}

\label{tab:abi_ref}

\end{table*}

\subsection{Decoder}

Our module has three decoders: trajectory decoder, probability decoder, and quality decoder. We represent them all as 
 MLP networks. We use the following model hyperparameters:
 \begin{itemize}[leftmargin=2em]
    \item Number of layers: 2
     \item Hidden size: 64
     \item Output size: $2\cdot{T_f}$ for trajectory decoder, and 1 for probability/quality decoder.
 \end{itemize}

\subsection{Optimization}

We train our model to minimize the negative log-likelihood of the ground truth trajectory. The training hyperparameters are set as follows:
\begin{itemize}[leftmargin=2em]
    \item Loss balance constant $\alpha$ in Sec.~\ref{sec:train}: 0.01
    \item Number of refinement iteration $I$: 5
    \item Number of training epochs: 32
    \item Batch size: 8 for one single GPU and we use 8 GPUs
    \item Learning rate schedule: Cosine
    \item Initial learning rate: 0.001
    \item Optimizer: AdamW
    \item Weight decay: 0.0001
\end{itemize}

\section{Evaluation Details}
\label{sec:eva}
\subsection{Standard Metric}
\noindent \textbf{minADE \& minFDE.} minADE measures the Euclidean
distance error averaged over all timesteps for the closest
prediction, relative to ground truth. In contrast, minFDE
considers only the distance error at the final timestep. 
\\

\noindent \textbf{Miss Rate.} Miss rate is a
measure of what fraction of scenarios fail to generate any
predictions within the lateral and longitudinal error thresholds, relative to the ground truth future. In Argoverse and Argoverse 2 dataset, the threshold is set to 2.0 by default.

\subsection{Additional Metric}

\noindent \textbf{\#Param. Analysis.} \#Param. is the total number of model parameters. It reveals the model's memory cost. We count the model's parameters via PyTorch Lightning.
\\

\noindent \textbf{Flops Analysis.} Flops specifically refer to the number of floating-point operations when running the model, such as the matrix multiplications and activations. Flops are often used to measure the computational cost or complexity of a model. We measure Flops of model forward with batch\_size=1 during inference so gradient calculations are not considered.
\\

\noindent \textbf{Latency Analysis.} Latency is the time required to execute the model, which is generally used in the context of measuring computation efficiency. To simulate realistic settings where multiple agents exist, we measure the latency of the model with batch\_size=32 over Argoverse / Argoveres 2 val set.

\begin{figure*}[t!]
\centering
\begin{minipage}[t]{0.33\textwidth}
\centering
\includegraphics[width=6cm]{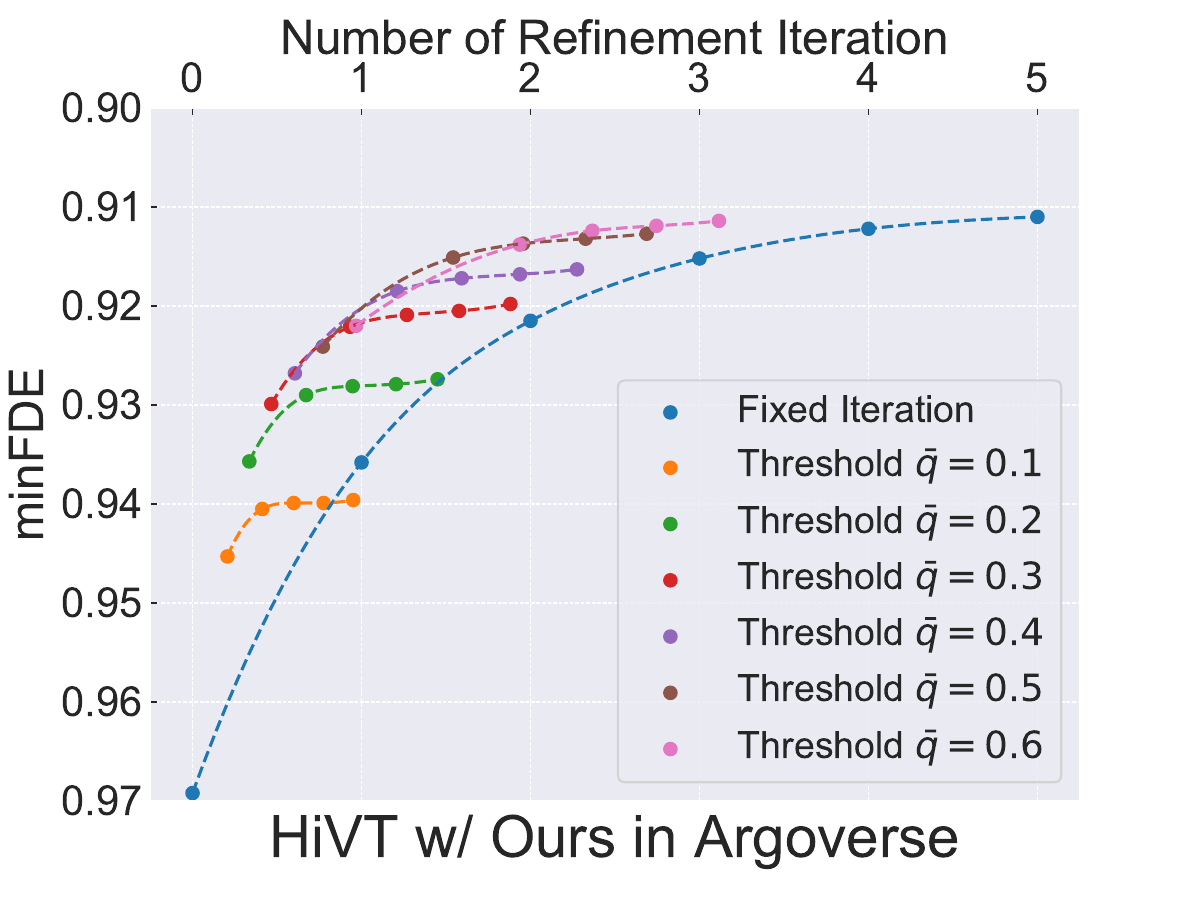}
\end{minipage}
    \vspace{-3mm}
\begin{minipage}[t]{0.33\textwidth}
\centering
\includegraphics[width=6cm]{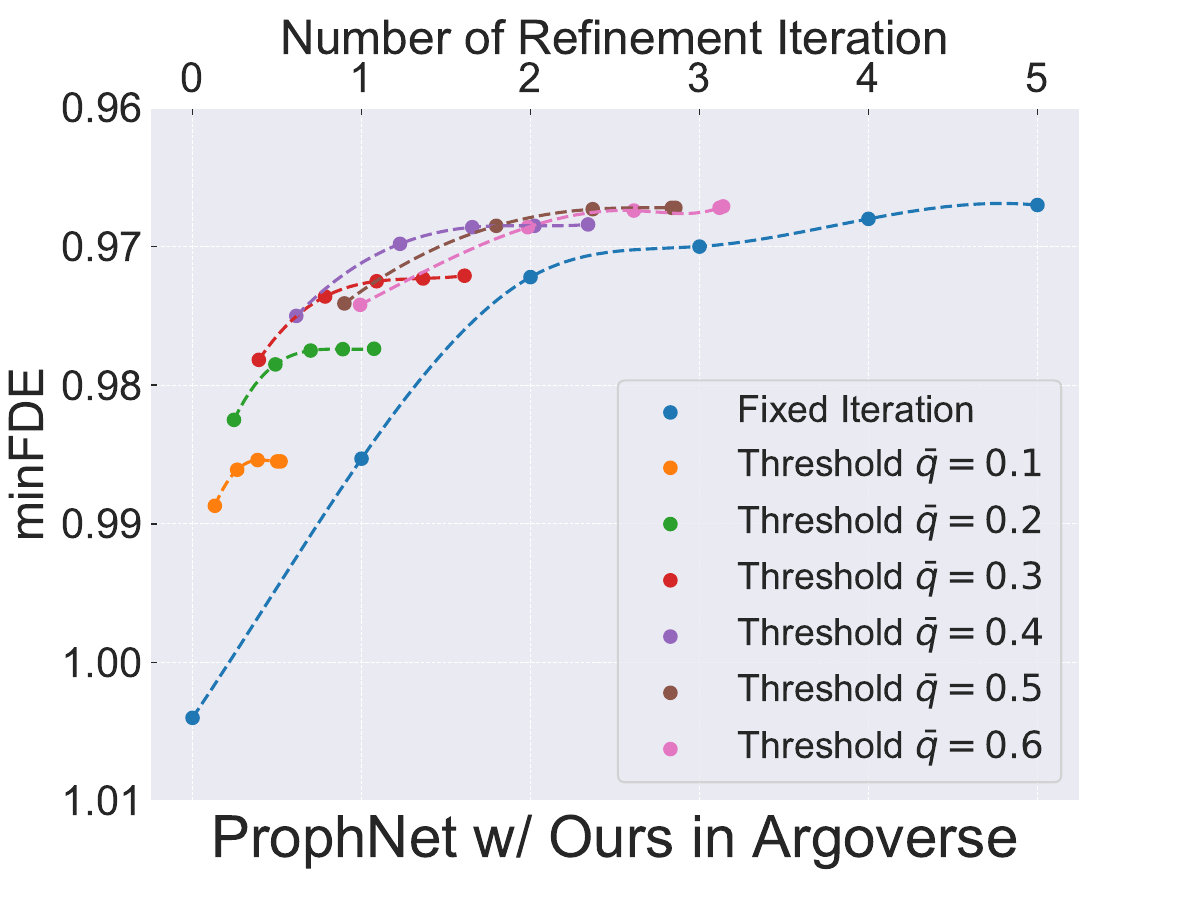}
\end{minipage}
\begin{minipage}[t]{0.33\textwidth}
\centering
\includegraphics[width=6cm]{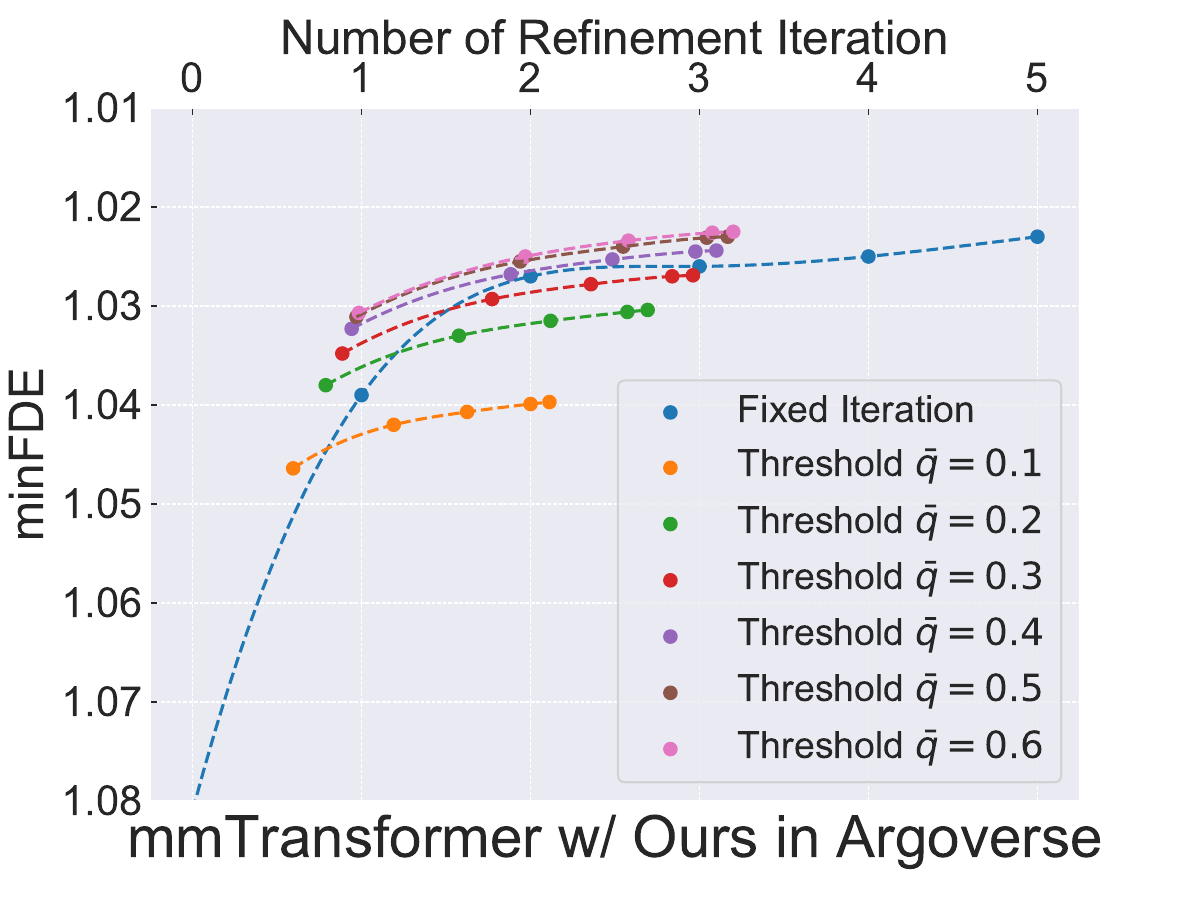}
\end{minipage}
\vfill
\vspace{6mm}
\begin{minipage}[t]{0.33\textwidth}
\includegraphics[width=6cm]{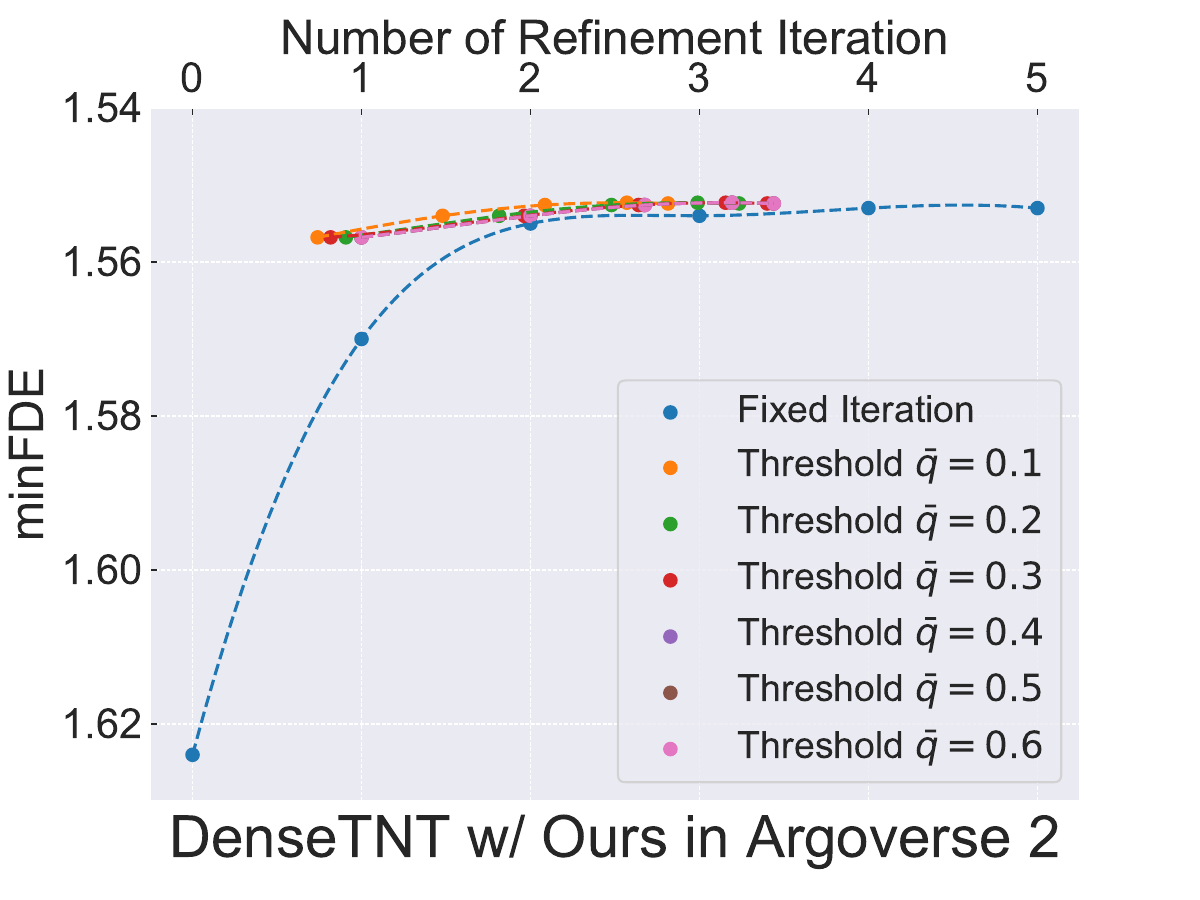}
\end{minipage}
    \vspace{-3mm}
\begin{minipage}[t]{0.33\textwidth}
\centering
\includegraphics[width=6cm]{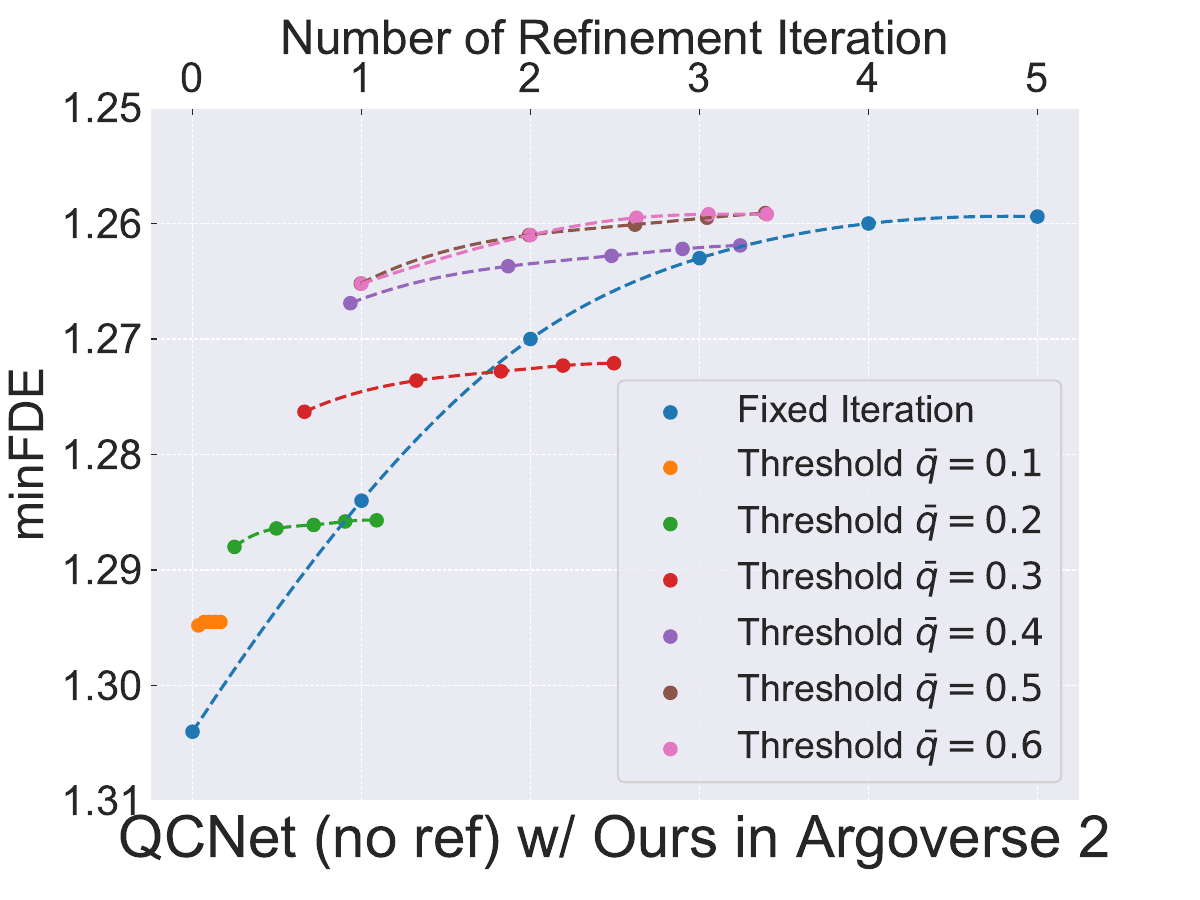}
\end{minipage}
\begin{minipage}[t]{0.33\textwidth}
\centering
\includegraphics[width=6cm]{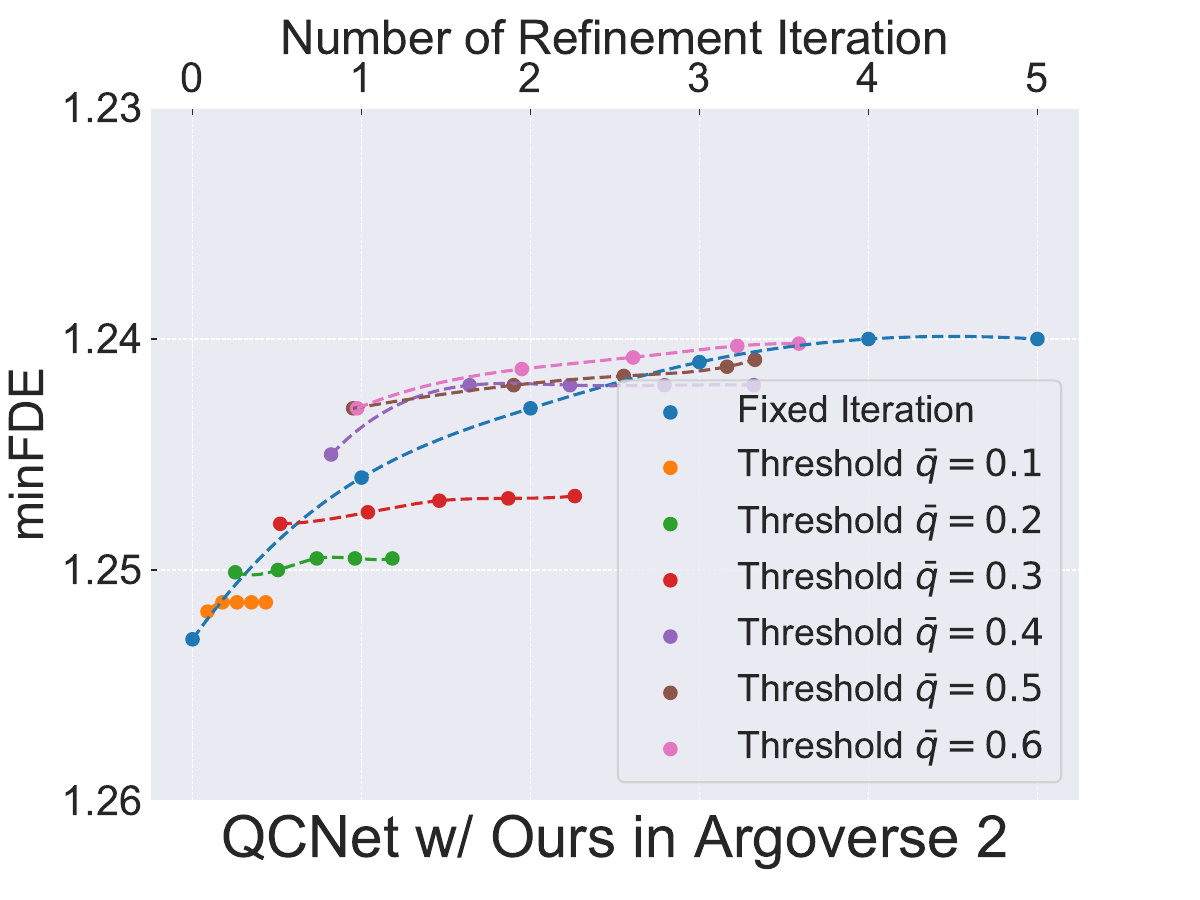}
\end{minipage}
\caption{Comparison between the fixed and adaptive number of refinement iterations, when we apply our SmartRefine on all six considered backbones respectively (on val set of Argoverse and Argoverse 2). The blue curve represents fixed refinement iterations (for both training and inference).
Other curves denote adaptive refinement iterations with different quality score threshold $\Bar{q}$ during inference time (5 refinement iterations are utilized during training). For each threshold, we ablate different limits for maximum refinement iteration $I'$, resulting in 5 points for each curve (refer to Algorithm~\ref{algo:inference} for detailed descriptions of $\Bar{q}$ and $I'$).
Two observations: 1) on HiVT, ProphNet, DenseTNT, and QCNet (no ref), our adaptive refinement strategy outperforms the fixed refinement strategy given any threshold $\bar{q}$. 2) on mmTransformer, and QCNet, our adaptive refinement strategy outperforms the fixed refinement strategy when we set a higher threshold $\bar{q}$ (0.4, 0.5, 0.6) when we decide whether another refinement iteration is needed.}
\label{fig:adap_all_argo1}
\vspace{-1em}
\end{figure*}

\section{Performance on Argoverse 2 Leaderboard}
\label{sec:lead}

In Table~\ref{tab:lead}, we show the details of how our SmartRefine performs on the Argoverse 2 leaderboard (single agent track). At the time of the paper submission, our SmartRefine with QCNet as trajectory generation backbone ranked \#8 on the leaderboard. Methods before ours are all unpublished except two methods: 
\begin{itemize}
    \item The \#3 method QCNet is the ensemble version of QCNet, while our method does not utilize ensemble. For a fair comparison, our method outperforms the ensemble-free version of QCNet (as marked in the second last row). 
    \item The \#5 method is linked to GANet~\citep{wang2023ganet}, which was published before.
    However, the performance in the original published GANet paper (as marked in the last row) is much poorer than that of the \#5 method.
    We suppose the current \#5 performance is obtained by the ensemble version of GANet or an unpublished extended version of GANet.
\end{itemize}
Thus our method outperforms all published ensemble-free works on the Argoverse 2 leaderboard (single agent track) at the time of the paper submission.

\section{Comparison with other refinement methods} 
\label{sec:abi_ref}

In the refinement methods listed in Table~\ref{tab:refinement summary}, DCMS and R-Pred are not open-sourced, and MTR is intended for Waymo dataset. For a fair and convenient comparison, we follow their ideologies to reproduce them under our framework. As in Table~\ref{tab:abi_ref}, DCMS has the worst minFDE but the best latency since it doesn't use any contexts. QCNet performs well but has the largest latency in Argoverse 2 with big maps, as it unselectively utilizes all the context. R-Pred selectively uses context along the whole trajectory and achieves good performance compared to other one-iteration methods (DCMS and QCNet). MTR as a multi-iteration refinement method outperforms one-iteration methods but has higher latency. Our method which utilizes adaptive refinement achieves the best trade-off in minFDE and latency.

\begin{table*}[!t]
\begin{minipage}{0.3\linewidth}
    \centering
        \resizebox{\linewidth}{!}{
    \begin{tabular}{c|cc}
        \toprule
            \multicolumn{1}{c|}{\multirow{3}{*}{\#Anchor numbers}} & \multicolumn{2}{c}{\multirow{2}{*}{\tabincell{c}{HiVT w/ Ours\\Argoverse}}} \\ 
            & & \\
            \cmidrule(r){2-3}
            & minFDE & \#Param.      \\ \midrule
         1 & 0.928 & 134K \\
         
         \cellcolor{lightgrey}2 & \cellcolor{lightgrey}0.911 & \cellcolor{lightgrey}207K\\
         3 & 0.911 & 280K\\
         5 & 0.915 & 433K \\
         6 & 0.916 & 509K  \\
         \bottomrule
    \end{tabular}
    }
    \vspace{-1em}
    \vspace{-1em}
\end{minipage}\hfill
\begin{minipage}{0.3\linewidth}
    \centering
        \resizebox{\linewidth}{!}{
    \begin{tabular}{c|cc}
        \toprule
            \multicolumn{1}{c|}{\multirow{3}{*}{\#Anchor numbers}} & \multicolumn{2}{c}{\multirow{2}{*}{\tabincell{c}{ProphNet w/ Ours\\Argoverse}}} \\ 
            & & \\ 
            \cmidrule(r){2-3}
            & minFDE & \#Param.      \\ \midrule
         1 & 0.978 & 143K \\
         
         \cellcolor{lightgrey}2 & \cellcolor{lightgrey}0.967 & \cellcolor{lightgrey}216K\\
         3 & 0.967 & 290K\\
         5 & 0.966 & 442K \\
         6 & 0.967 & 518K  \\
         \bottomrule
    \end{tabular}
    }
    \vspace{-1em}
    \vspace{-1em}
\end{minipage}\hfill
\begin{minipage}{0.33\linewidth}
    \centering
        \resizebox{\linewidth}{!}{
    \begin{tabular}{c|cc}
        \toprule
            \multicolumn{1}{c|}{\multirow{3}{*}{\#Anchor numbers}} & \multicolumn{2}{c}{\multirow{2}{*}{\tabincell{c}{mmTransformer w/ Ours\\Argoverse}}} \\ 
            & & \\
            \cmidrule(r){2-3}
            & minFDE & \#Param.      \\ \midrule
         1 & 1.035 & 143K \\
         
         \cellcolor{lightgrey}2 & \cellcolor{lightgrey}1.023 & \cellcolor{lightgrey}216K\\
         3 & 1.021 & 290K\\
         5 & 1.020 & 442K \\
         6 & 1.018 & 518K  \\
         \bottomrule
    \end{tabular}
    }
    \vspace{-1em}
    \vspace{-1em}
\end{minipage}
\vfill
\vspace{11mm}

\begin{minipage}{0.31\linewidth}
    \centering
        \resizebox{\linewidth}{!}{
    \begin{tabular}{c|cc}
        \toprule
            \multicolumn{1}{c|}{\multirow{3}{*}{\#Anchor numbers}} & \multicolumn{2}{c}{\multirow{2}{*}{\tabincell{c}{DenseTNT w/ Ours\\Argoverse}}} \\
            & & \\
            \cmidrule(r){2-3}
            & minFDE & \#Param.      \\ \midrule
         1 & 1.582 & 176K \\
         3 & 1.555 & 327K \\
         
         \cellcolor{lightgrey}4 & \cellcolor{lightgrey}1.553 & \cellcolor{lightgrey}406K\\
         5 & 1.552 & 486K \\
         6 & 1.553 & 566K \\
         \bottomrule
    \end{tabular}
    }
    \vspace{-1em}
    \vspace{-1em}
\end{minipage}\hfill
\begin{minipage}{0.33\linewidth}
    \centering
        \resizebox{\linewidth}{!}{
    \begin{tabular}{c|cc}
        \toprule
            \multicolumn{1}{c|}{\multirow{3}{*}{\#Anchor numbers}} & \multicolumn{2}{c}{\multirow{2}{*}{\tabincell{c}{QCNet (no ref) w/ Ours\\Argoverse}}} \\ 
            & & \\
            \cmidrule(r){2-3}
            & minFDE & \#Param.      \\ \midrule
         1 & 1.282 & 142K \\
         3 & 1.260 & 284K \\
         
         \cellcolor{lightgrey}4 & \cellcolor{lightgrey}1.258 & \cellcolor{lightgrey}359K \\
         5 & 1.258 & 435K \\
         6 & 1.259 & 511K \\
         \bottomrule
    \end{tabular}
    }
    \vspace{-1em}
    \vspace{-1em}
\end{minipage}\hfill
\begin{minipage}{0.30\linewidth}
    \centering
        \resizebox{\linewidth}{!}{
    \begin{tabular}{c|cc}
        \toprule
            \multicolumn{1}{c|}{\multirow{3}{*}{\#Anchor numbers}} & \multicolumn{2}{c}{\multirow{2}{*}{\tabincell{c}{QCNet w/ Ours\\Argoverse}}} \\ 
            & & \\
            \cmidrule(r){2-3}
            & minFDE & \#Param.      \\ \midrule
         1 & 1.245 & 142K \\
         3 & 1.242 & 284K \\
         
         \cellcolor{lightgrey}4 & \cellcolor{lightgrey}1.240 & \cellcolor{lightgrey}359K  \\
         5 & 1.242 & 435K \\
         6 & 1.241 & 511K \\
         \bottomrule
    \end{tabular}
    }
    \vspace{-1em}
    \vspace{-1em}
\end{minipage}

\vspace{7mm}
\caption{Ablation study on the number of anchors, when we apply our SmartRefine on all six considered backbones respectively (on val set of Argoverse and Argoverse 2). 
We can see a common trend that increasing the anchor number reduces the minFDE. However, excessively increasing the number of anchors is ineffective, as it brings much larger model parameters with the same or slightly worse accuracy. 
Also, the two datasets desire different numbers of anchors because they consider different lengths of the prediction horizon.
Thus we set the anchor number as \textbf{2} for Argoverse (marked grey), and \textbf{4} for Argoverse 2 experiments (marked grey). The experiments reported in the main paper are based on these two settings.}
\label{tab:all_anchor_argo1_2}
\end{table*}

\section{Ablation Studies with All Backbones}
\label{sec:abi_all}
In this main paper, we only reported ablation studies on certain backbones due to the page limit. Here we show the ablation studies on all backbones.
\subsection{Refinement Iterations}
In Fig.~\ref{fig:adaptive} of the main paper, we compared the fixed refinement strategy and our proposed adaptive strategy on two backbones: HiVT (Argoverse), and QCNet no refine (Argoverse 2). Here we show the results of our SmartRefine with all six considered backbones, shown in Fig.~\ref{fig:adap_all_argo1}. Specifically, the blue curve denotes refinement with a fixed number of iterations (for both training and inference). 
Other curves denote adaptive refinement iterations with different quality score threshold $\Bar{q}$ during inference (5 refinement iterations are utilized during training). For each threshold, we ablate different limits for maximum refinement iteration $I'$, resulting in 5 points for each curve. Readers are referred to Algorithm~\ref{algo:inference} for detailed descriptions of $\Bar{q}$ and $I'$.
As shown in Fig.~\ref{fig:adap_all_argo1}, the performance of all backbones can be improved by refinement. When we compare our adaptive strategy with the fixed strategy: 1) on HiVT, ProphNet, DenseTNT, and QCNet (no ref), our adaptive refinement strategy outperforms the fixed refinement strategy basically given any threshold $\bar{q}$. 2) on mmTransformer, and QCNet, our adaptive refinement strategy outperforms the fixed refinement strategy when we set a higher threshold $\bar{q}$ (0.4, 0.5, 0.6) when we decide whether another refinement iteration is needed.

\subsection{Anchor Numbers}

The results are shown in Table~\ref{tab:all_anchor_argo1_2}.
We can see a common trend that increasing the anchor number reduces the minFDE. However, excessively increasing the number of anchors is ineffective, as it brings much larger model parameters with the same or slightly worse accuracy. 
Also, the two datasets desire different numbers of anchors because they consider different lengths of the prediction horizon.
Thus we set the anchor number as \textbf{2} for Argoverse (marked grey), and \textbf{4} for Argoverse 2 (marked grey). The experiments reported in the main paper are based on these two settings.

\subsection{Context Representation}

The results are shown in Table~\ref{tab:all_context_argo1_2}. We can see our adaptive anchor-centric encoding effectively outperforms the fixed agent-centric context encoding, on all backbones.

\subsection{Retrieval Radius}

The results are shown in Table~\ref{tab:all_radius_argo1_2}.
Common observations can be drawn: 1) the fixed retrieval radius from 50 to 2 can be sub-optimal, as a large retrieval radius might lead to redundant or irrelevant context information, while a small radius might not provide sufficient context for refinement.
2) our SmartRefine achieves lower prediction error and Flops, by adapting the radius to each agent's velocity, and decaying the radius with the number of refinement iterations (see details in Sec~\ref{sec:anchor&radius}). Here we compare two strategies for radius decay: linear decay and exponential decay. We adopt exponential decay as it outperforms linear decay.

\section{Quality Score of All Backbones}
\label{sec:score_all}
\begin{figure*}[!t]
    \centering
    \begin{minipage}{\linewidth}{
    \centering
    \includegraphics[width=0.98\textwidth]{figure/train_dis_test_1113_crop.pdf}
    \vspace{-3mm}
    \captionsetup{labelformat=empty}
    \caption{HiVT w/ Ours in Argoverse}
     \centering
    \includegraphics[width=0.98\textwidth]{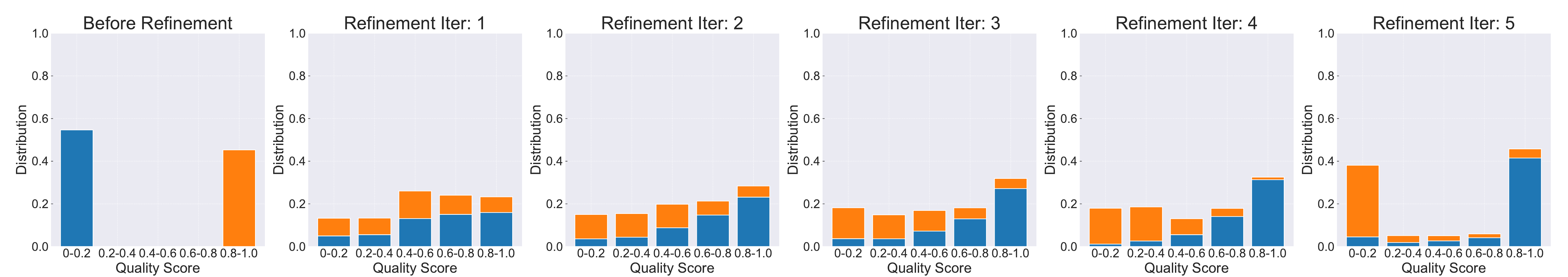}
    \vspace{-3mm}
    \captionsetup{labelformat=empty}
    \caption{ProphNet w/ Ours in Argoverse}
        \centering
    \includegraphics[width=0.98\textwidth]{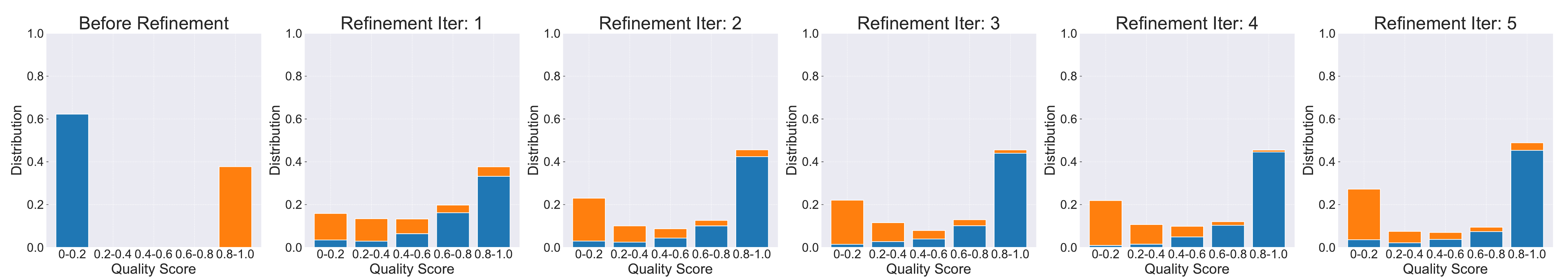}
    \vspace{-3mm}
    \captionsetup{labelformat=empty}
    \caption{ mmTransformer w/ Ours in Argoverse}
    \centering
    \includegraphics[width=0.98\textwidth]{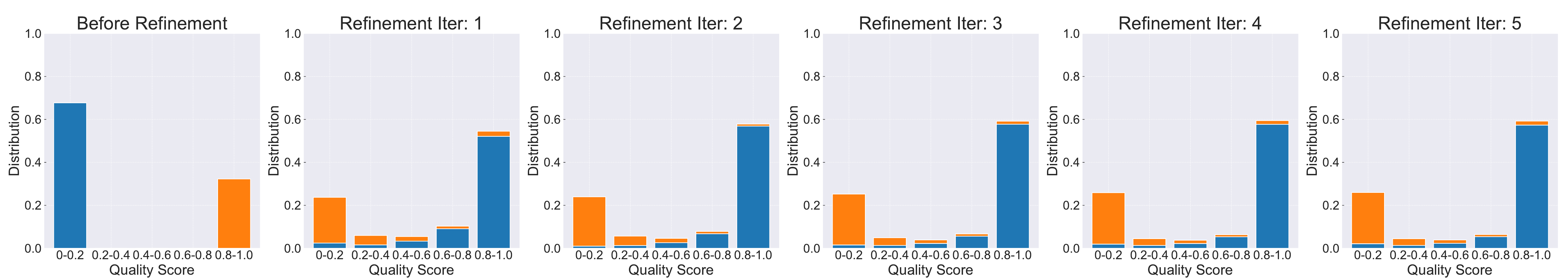}
    \vspace{-3mm}
    \captionsetup{labelformat=empty}
    \caption{DenseTNT w/ Ours in Argoverse 2}
     \centering
    \includegraphics[width=0.98\textwidth]{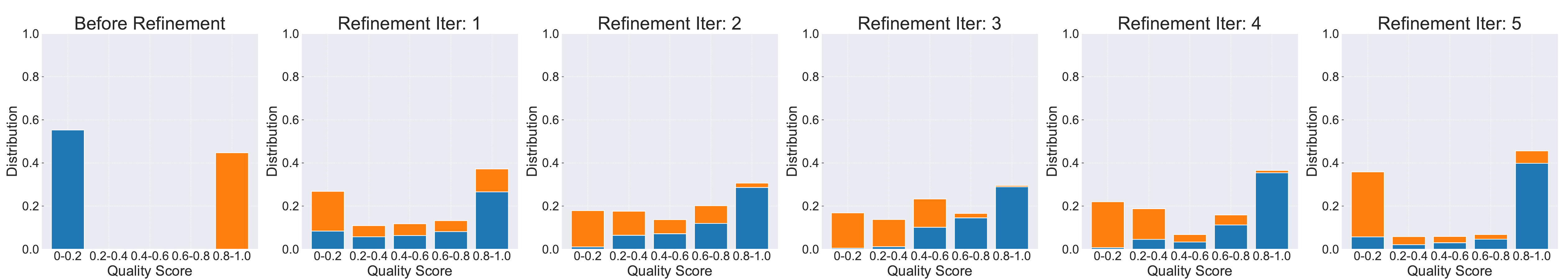}
    \vspace{-3mm}
    \captionsetup{labelformat=empty}
    \caption{QCNet (no ref) w/ Ours in Argoverse 2}
        \centering
    \includegraphics[width=0.98\textwidth]{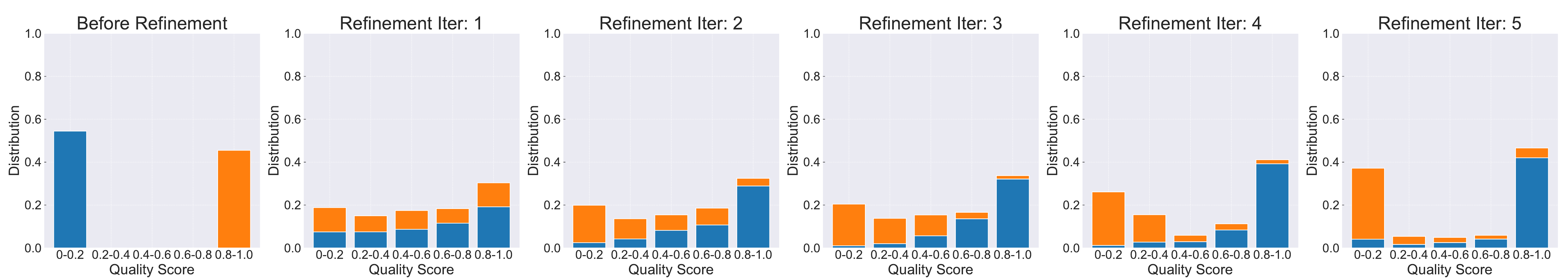}
    \vspace{-3mm}
    \captionsetup{labelformat=empty}
    \caption{QCNet w/ Ours in Argoverse 2}
    }
    \end{minipage}
    \vspace{-3mm}
    \caption{The quality score distribution over multiple refinement iterations, when we apply our SmartRefine on all six considered backbones respectively (Argoverse and Argoverse 2 training set).
    Specifically, for each predicted trajectory, we measure its accuracy using the quality score. We will track how the quality score changes along the multi-iteration refinements.
    Common observations on the six backbones can be drawn: \textit{1) not every trajectory benefits from refinement;} \textit{2) the overall performance becomes better after refinement.} These results demonstrate the necessity of adaptive refinement.
    }
    \label{fig:all_score_argo1_2}
\end{figure*}

The quality score distribution over multiple refinement iterations, when we apply our SmartRefine on all six considered backbones respectively. Results are shown in Fig.~\ref{fig:all_score_argo1_2}.
Specifically, for each predicted trajectory, we measure its accuracy using the quality score. We will track how the quality score changes along the multi-iteration refinements.

Common observations on the six backbones can be drawn: \textit{1) not every trajectory benefits from refinement;} \textit{2) the overall performance becomes better after refinement.} These results demonstrate the necessity of adaptive refinement.

\section{Visualization of Refinement}
\label{sec:vis}
\begin{figure*}[!h]
    \centering
    \includegraphics[width=1\linewidth]{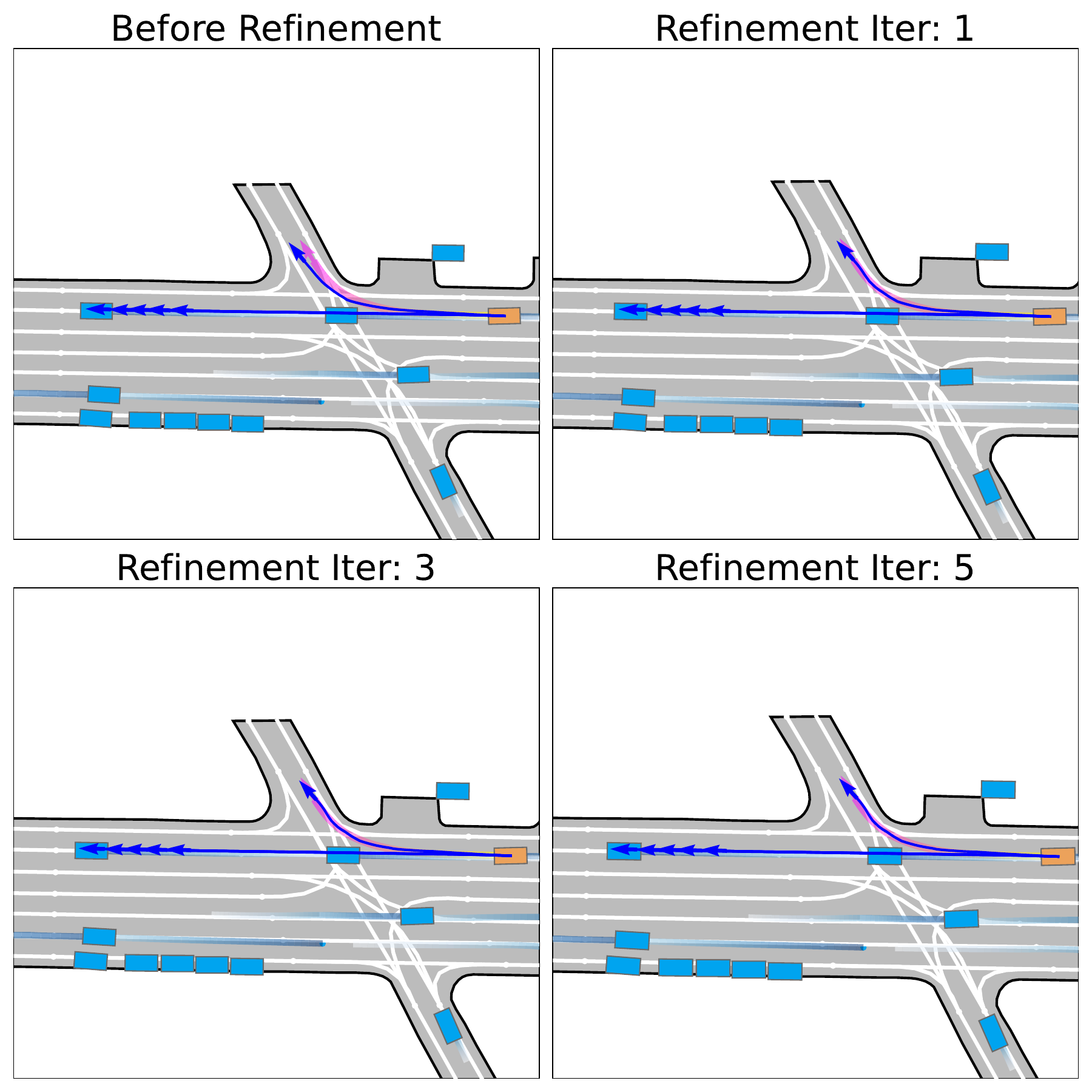}
    \caption{Visualization results. The dark blue arrows are multi-nodal predictions of the agent by model and the pink arrow is the ground truth future trajectory respectively. The trajectory (turn right) gets closer to the ground truth after refinement.}
    \label{fig:vis_1}
    \vspace{-1em}
\end{figure*}

\begin{figure*}[!h]
    \centering
    \includegraphics[width=1\linewidth]{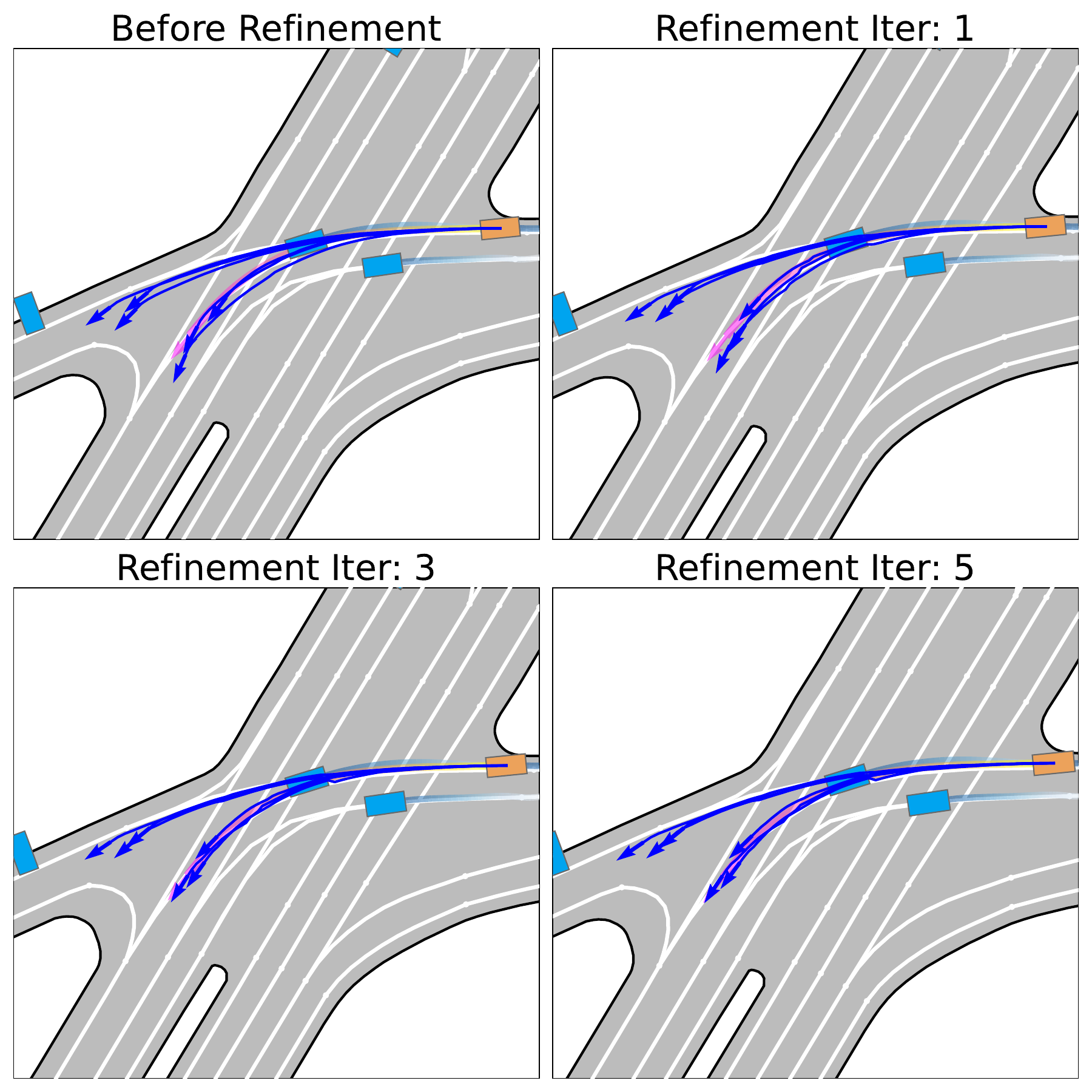}
    \caption{Visualization results.  The dark blue arrows are multi-nodal predictions of the agent by model and the pink arrow is the ground truth future trajectory respectively. The trajectory closest to the ground truth gets closer after refinement.}
    \label{fig:vis_5}
    \vspace{-1em}
\end{figure*}

\begin{figure*}[!h]
    \centering
    \includegraphics[width=1\linewidth]{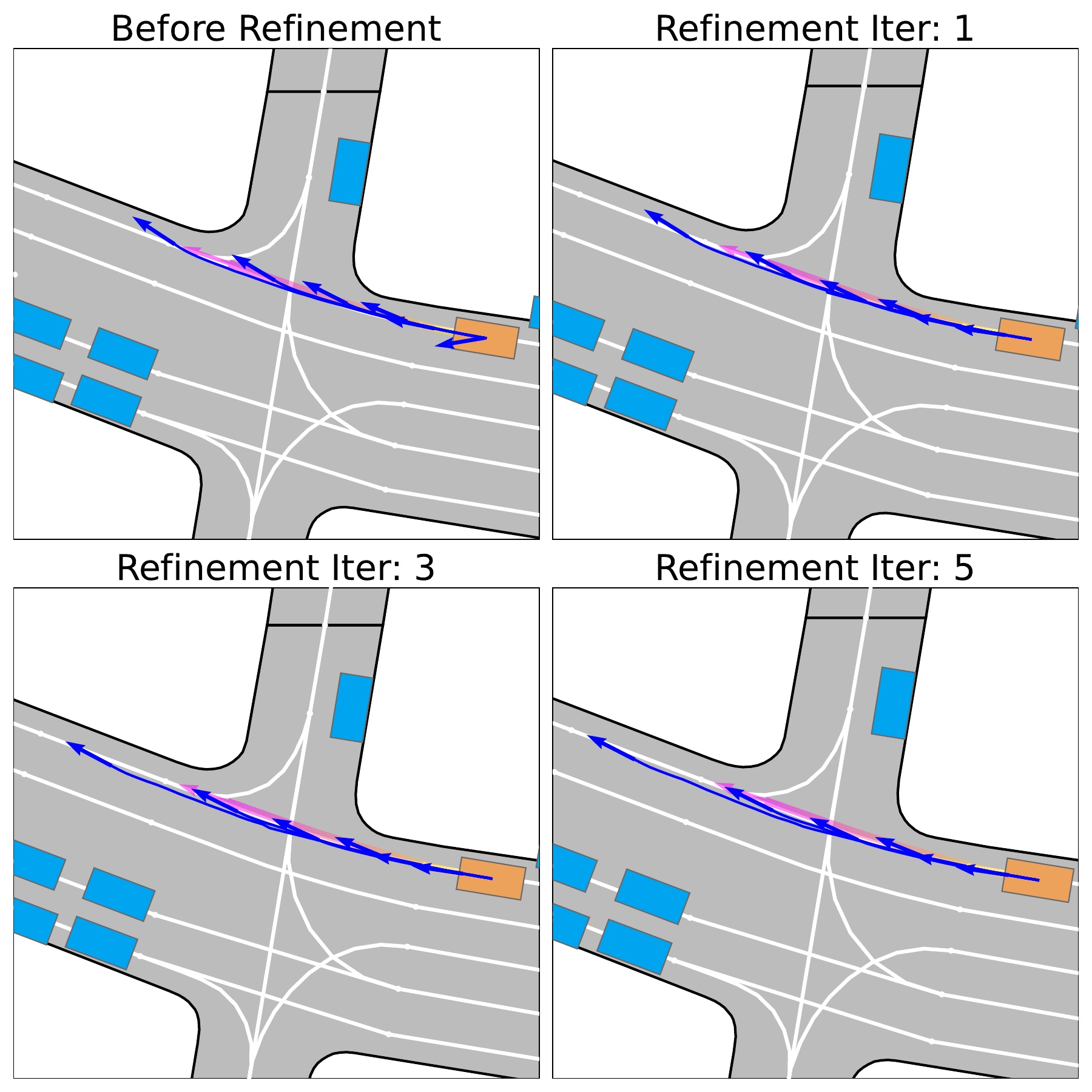}
    \caption{Visualization results.  The dark blue arrows are multi-nodal predictions of the agent by model and the pink arrow is the ground truth future trajectory respectively. The shortest trajectory gets more aligned toward the ground truth direction, and the trajectory closest to the ground truth gets closer after refinement.}
    \label{fig:vis_2}
    \vspace{-1em}
\end{figure*}

\begin{figure*}[!h]
    \centering
    \includegraphics[width=1\linewidth]{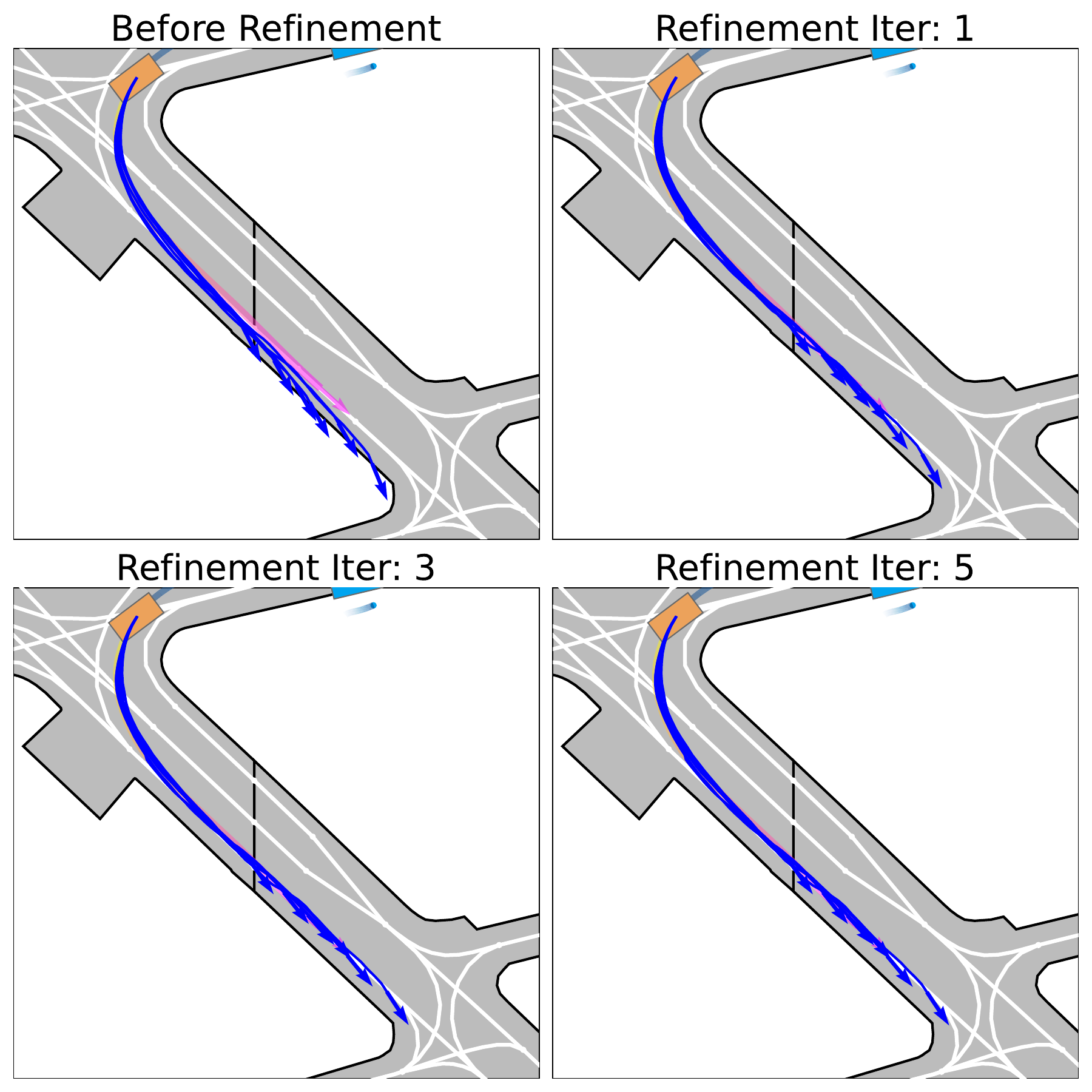}
    \caption{Visualization results. The dark blue arrows are multi-nodal predictions of the agent by model and the pink arrow is the ground truth future trajectory respectively. All trajectories get closer to the ground truth after refinement.}
    \label{fig:vis_3}
    \vspace{-1em}
\end{figure*}

\begin{figure*}[!h]
    \centering
    \includegraphics[width=1\linewidth]{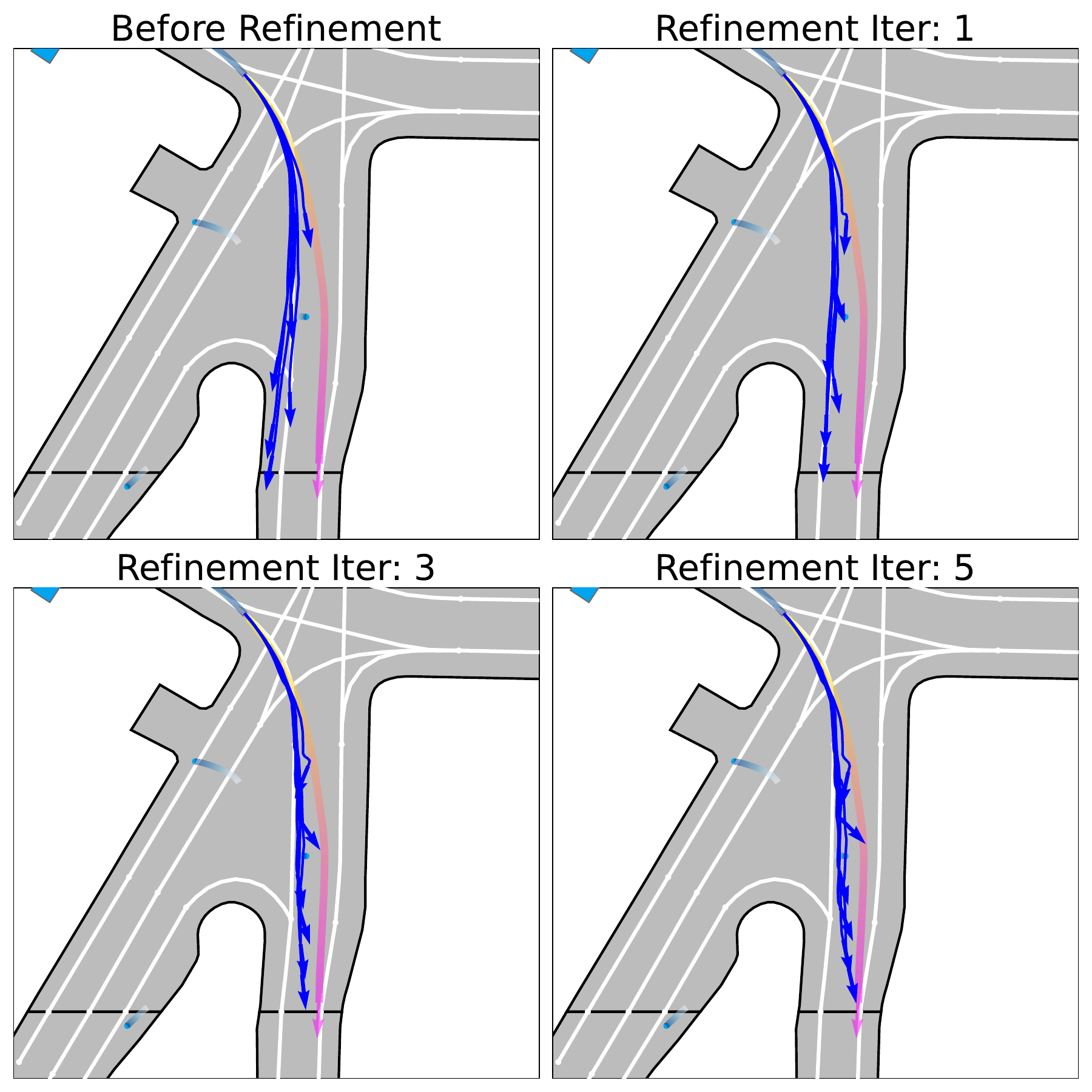}
    \caption{Visualization results when the future trajectory of a pedestrian is predicted.  The dark blue arrows are multi-nodal predictions of the agent by model and the pink arrow is the ground truth future trajectory respectively. All trajectories get closer to the ground truth after refinement.}
    \label{fig:vis_4}
    \vspace{-1em}
\end{figure*}

As shown in Fig.~\ref{fig:vis_1}, Fig.~\ref{fig:vis_5}, Fig.~\ref{fig:vis_2}, Fig.~\ref{fig:vis_3} and Fig.~\ref{fig:vis_4}, we show the visualization results of 
the predicted trajectories by our method, before and after refinement. These results demonstrate how our method can refine the trajectory to be closer to ground truth and be more compliant to the road context.

\begin{table*}[!t]
\begin{minipage}{0.29\linewidth}
        \centering
    \resizebox{\linewidth}{!}{
    \begin{tabular}{c|c}
        \toprule
          \multirow{3}{*}{Context Encoding}  & 
          \multirow{2}{*}{\tabincell{c}{HiVT w/ Ours\\Argoverse}} \\
            & \\
          \cmidrule(r){2-2} 
          & minFDE      \\ \midrule
         Agent-Centric & 0.941  \\
         Anchor-Centric & 0.911 \\
         \bottomrule
    \end{tabular}
    }
    \vspace{-1em}
\end{minipage}\hfill
\begin{minipage}{0.30\linewidth}
        \centering
    \resizebox{\linewidth}{!}{
    \begin{tabular}{c|c}
        \toprule
          \multirow{3}{*}{Context Encoding}  & 
          \multirow{2}{*}{\tabincell{c}{ProphNet w/ Ours\\Argoverse}} \\
            & \\
          \cmidrule(r){2-2} 
          & minFDE      \\ \midrule
         Agent-Centric & 0.988  \\
         Anchor-Centric & 0.967 \\
         \bottomrule
    \end{tabular}
    }
    \vspace{-1em}
\end{minipage}\hfill
\begin{minipage}{0.33\linewidth}
        \centering
    \resizebox{\linewidth}{!}{
    \begin{tabular}{c|c}
        \toprule
          \multirow{3}{*}{Context Encoding}  & 
          \multirow{2}{*}{\tabincell{c}{mmTransformer w/ Ours\\Argoverse}} \\
            & \\
          \cmidrule(r){2-2} 
          & minFDE      \\ \midrule
         Agent-Centric & 1.064  \\
         Anchor-Centric & 1.023 \\
         \bottomrule
    \end{tabular}
    }
    \vspace{-1em}
\end{minipage}
\vfill
\vspace{7mm}
\begin{minipage}{0.31\linewidth}
        \centering
    \resizebox{\linewidth}{!}{
    \begin{tabular}{c|c}
        \toprule
          \multirow{3}{*}{Context Encoding}  & 
          \multirow{2}{*}{\tabincell{c}{DenseTNT w/ Ours\\Argoverse}} \\
            & \\
          \cmidrule(r){2-2} 
          & minFDE      \\ \midrule
         Agent-Centric & 1.589  \\
         Anchor-Centric & 1.553 \\
         \bottomrule
    \end{tabular}
    }
    \vspace{-1em}
\end{minipage}\hfill
\begin{minipage}{0.33\linewidth}
        \centering
    \resizebox{\linewidth}{!}{
    \begin{tabular}{c|c}
        \toprule
          \multirow{3}{*}{Context Encoding}  & 
          \multirow{2}{*}{\tabincell{c}{QCNet (no ref) w/ Ours\\Argoverse}} \\
            & \\
          \cmidrule(r){2-2} 
          & minFDE      \\ \midrule
         Agent-Centric & 1.276  \\
         Anchor-Centric & 1.258 \\
         \bottomrule
    \end{tabular}
    }
    \vspace{-1em}
\end{minipage}\hfill
\begin{minipage}{0.3\linewidth}
        \centering
    \resizebox{\linewidth}{!}{
    \begin{tabular}{c|c}
        \toprule
          \multirow{3}{*}{Context Encoding}  & 
          \multirow{2}{*}{\tabincell{c}{QCNet w/ Ours\\Argoverse}} \\
            & \\
          \cmidrule(r){2-2} 
          & minFDE      \\ \midrule
         Agent-Centric &  1.249 \\
         Anchor-Centric & 1.240  \\
         \bottomrule
    \end{tabular}
    }
    \vspace{-1em}
\end{minipage}
\vspace{3mm}
\caption{Ablation study on how the contexts are encoded,
when we apply our SmartRefine on all six considered backbones respectively (on val set of Argoverse and Argoverse 2). We can see our adaptive anchor-centric encoding effectively outperforms the fixed agent-centric context encoding, on all backbones.}
\label{tab:all_context_argo1_2}
\end{table*}

\begin{table*}[!t]
\begin{minipage}{0.32\linewidth}
        \centering
    \resizebox{\linewidth}{!}{
    \begin{tabular}{c|c|c|c}
          \toprule
          & \multicolumn{3}{c}{HiVT w/ Ours (Argoverse)} \\
          \cmidrule(r){2-4}
          & Retrieval Radius   & minFDE &Flops (M)      \\ \midrule
         \multirow{4}{*}{\tabincell{c}{Fixed \\Radius}} 
         &50 & 0.926&2,297  \\
         &20 & 0.923& 722 \\
         &10 & 0.921&325 \\
         &2 & 0.930&58 \\
         \midrule
         \multirow{2}{*}{\tabincell{c}{Adaptive\\Radius}} 
         &$R_{max}$=10, $R_{min}$=2, linear & 0.911&245 \\
         &$R_{max}$=10, $R_{min}$=2, exp~~~ & 0.911&130\\
         \bottomrule
    \end{tabular}
    }
    \vspace{-1em}
\end{minipage}\hfill
\begin{minipage}{0.32\linewidth}
        \centering
    \resizebox{\linewidth}{!}{
    \begin{tabular}{c|c|c|c}
          \toprule
          & \multicolumn{3}{c}{ProphNet w/ Ours (Argoverse)} \\
          \cmidrule(r){2-4}
          & Retrieval Radius   & minFDE &Flops (M)      \\ \midrule
         \multirow{4}{*}{\tabincell{c}{Fixed \\Radius}} 
         &50 & 0.976& 2,181 \\
         &20 & 0.974& 621 \\
         &10 & 0.975& 332 \\
         &2 & 0.980& 41\\
         \midrule
         \multirow{2}{*}{\tabincell{c}{Adaptive\\ Radius}} 
         &$R_{max}$=10, $R_{min}$=2, linear & 0.970&140 \\
         &$R_{max}$=10, $R_{min}$=2, exp~~~ & 0.967&132\\
         \bottomrule
    \end{tabular}
    }
    \vspace{-1em}
\end{minipage}\hfill
\begin{minipage}{0.32\linewidth}
        \centering
    \resizebox{\linewidth}{!}{
    \begin{tabular}{c|c|c|c}
          \toprule
          & \multicolumn{3}{c}{mmTransformer w/ Ours (Argoverse)} \\
          \cmidrule(r){2-4}
          & Retrieval Radius   & minFDE &Flops (M)      \\ \midrule
         \multirow{4}{*}{\tabincell{c}{Fixed\\ Radius}} 
         &50 & 1.042& 1,466  \\
         &20 & 1.030& 468 \\
         &10 & 1.031& 193 \\
         &2 & 1.041& 30 \\
         \midrule
         \multirow{2}{*}{\tabincell{c}{Adaptive\\ Radius}} 
         &$R_{max}$=10, $R_{min}$=2, linear & 1.028& 156 \\
         &$R_{max}$=10, $R_{min}$=2, exp~~~ & 1.023& 101 \\
         \bottomrule
    \end{tabular}
    }
    \vspace{-1em}
\end{minipage}
\vfill
\vspace{7mm}
\begin{minipage}{0.32\linewidth}
        \centering
    \resizebox{\linewidth}{!}{
    \begin{tabular}{c|c|c|c}
          \toprule
          & \multicolumn{3}{c}{DenseTNT w/ Ours (Argoverse 2)} \\
          \cmidrule(r){2-4}
          & Retrieval Radius   & minFDE &Flops (M)      \\ \midrule
         \multirow{4}{*}{\tabincell{c}{Fixed\\ Radius}} 
         &50 & 1.566 & 4,018 \\
         &20 & 1.556 & 1,476 \\
         &10 & 1.558 & 839  \\
         &2 & 1.561 & 181 \\
         \midrule
         \multirow{2}{*}{\tabincell{c}{Adaptive\\ Radius}} 
         &$R_{max}$=10, $R_{min}$=2, linear & 1.555 & 541\\
         &$R_{max}$=10, $R_{min}$=2, exp~~~ & 1.553 & 396\\
         \bottomrule
    \end{tabular}
    }
    \vspace{-1em}
\end{minipage}\hfill
\begin{minipage}{0.32\linewidth}
        \centering
    \resizebox{\linewidth}{!}{
    \begin{tabular}{c|c|c|c}
          \toprule
          & \multicolumn{3}{c}{QCNet (no ref) w/ Ours (Argoverse 2)} \\
          \cmidrule(r){2-4}
          & Retrieval Radius   & minFDE &Flops (M)      \\ \midrule
         \multirow{4}{*}{\tabincell{c}{Fixed\\ Radius}} 
         &50 & 1.266 & 4,337  \\
         &20 & 1.258&  1,691\\
         &10 & 1.261 & 995  \\
         &2 & 1.270 & 199 \\
         \midrule
         \multirow{2}{*}{\tabincell{c}{Adaptive\\ Radius}} 
         &$R_{max}$=10, $R_{min}$=2, linear & 1.258 & 577 \\
         &$R_{max}$=10, $R_{min}$=2, exp~~~ & 1.258 & 408 \\
         \bottomrule
    \end{tabular}
    }
    \vspace{-1em}
\end{minipage}\hfill
\begin{minipage}{0.32\linewidth}
        \centering
    \resizebox{\linewidth}{!}{
    \begin{tabular}{c|c|c|c}
          \toprule
          & \multicolumn{3}{c}{QCNet w/ Ours (Argoverse 2)} \\
          \cmidrule(r){2-4}
          & Retrieval Radius   & minFDE &Flops (M)      \\ \midrule
         \multirow{4}{*}{\tabincell{c}{Fixed\\Radius}} 
         &50 & 1.244 & 4,560 \\
         &20 & 1.245 & 1,756  \\
         &10 & 1.243 & 869 \\
         &2 &  1.246 & 186 \\
         \midrule
         \multirow{2}{*}{\tabincell{c}{Adaptive\\Radius}} 
         &$R_{max}$=10, $R_{min}$=2, linear & 1.241 & 594 \\
         &$R_{max}$=10, $R_{min}$=2, exp~~~ & 1.240 & 410 \\
         \bottomrule
    \end{tabular}
    }
    \vspace{-1em}
\end{minipage}

\vspace{3mm}
\caption{Ablation study on the fixed and adaptive radius for context retrieval, when we apply our SmartRefine on all six considered backbones respectively (on val set of Argoverse and Argoverse 2). Common observations can be drawn: 1) the fixed retrieval radius from 50 to 2 can be sub-optimal, as a large retrieval radius might lead to redundant or irrelevant context information, while a small radius might not provide sufficient context for refinement.
2) our SmartRefine achieves lower accuracy and Flops, by adapting the radius to each agent's velocity, and decaying the radius with the number of refinement iterations (see details in Sec~\ref{sec:anchor&radius}). Here we compare two strategies for radius decay: linear decay and exponential decay. We adopt exponential decay as it outperforms linear decay.
}
\label{tab:all_radius_argo1_2}
\end{table*}

\clearpage

\clearpage
\clearpage
\clearpage
\clearpage
\clearpage

\end{document}